\documentclass[letterpaper]{article} 
\usepackage[draft]{aaai2027}    
\usepackage[hyphens]{url}            
\usepackage{graphicx}                
\usepackage{natbib}                  
\usepackage{caption}                 
\usepackage{booktabs}
\usepackage{amsmath}
\usepackage{tcolorbox}
\tcbuselibrary{breakable}

\title{The Tell-Tale Trace: Detecting Reasoning Failures in LLMs\\Using Chain-of-Thought Dynamics}

\author{
    Shashwat Sourav\textsuperscript{\rm 1},
    Aishwarya Balwani\textsuperscript{\rm 2}
}
\affiliations{
    \textsuperscript{\rm 1}Department of Physics,
    Washington University in St. Louis\\
    \textsuperscript{\rm 2}Department of Developmental Neurobiology,
    St. Jude Children's Research Hospital\\
    s.shashwat@wustl.edu, aishwarya.balwani@stjude.org
}

\begin{document}
\maketitle

\begin{abstract}

Chain-of-thought (CoT) reasoning as a form of test-time computation not only improves large language model (LLM) performance, but also provides an observable interface to the model's reasoning process.
Existing approaches that leverage verbalized CoTs to monitor reasoning correctness, however, largely evaluate the semantic correctness or consistency of individual intermediate steps, rather than how the reasoning process evolves across the trace.
As a result, failures distributed across the reasoning trajectory, rather than those localized to a single incorrect step, remain comparatively underexplored.
Furthermore, verbalized CoTs need not faithfully reflect the model's internal reasoning, motivating analyses that do not treat individual statements as literal accounts of internal computation.
In this work, we therefore ask whether the dynamics of visible CoT can be leveraged to systematically distinguish successful from failed reasoning without assuming such semantic faithfulness, provide warning before answer emission, and guide targeted correction.
We study a range of LLMs on verifiable Boolean satisfiability tasks with variable complexity, enabling controlled comparisons near each model’s capability frontier.
Tagging CoT sentences by reasoning function reveals premature verification collapse on SAT problems: incorrect traces enter clause checking earlier, repeat similar operations, and finalize sooner.
On UNSAT problems, models presumptuously move towards incorrect SAT conclusions, checking candidate assignments rather than deriving contradictions across constructed cases.
Subsequently, a targeted proof-search prompt intervention raises Llama3-70B accuracy from 13.3\% to 85\%, correcting 84.6\% of these errors.
These results consequently show that capability failures can manifest as distributed, task-dependent changes in the sequencing, repetition, and timing of visible reasoning, and that CoT dynamics agnostic to whether the verbalized trace reflects the model’s internal computations can help diagnose and correct failures near the reasoning frontier.
\end{abstract}

\section{Introduction}

Chain-of-thought (CoT) reasoning \cite{wei2022chain} is a vital component of contemporary large language models (LLMs), driving their ability to solve intricate, multi-step tasks \cite{jaech2024openai, yang2025qwen3}.
By rewarding models to think out loud during post-training, LLMs are incentivized to optimize their intermediate reasoning in token space, resulting in natural language responses of the form $\langle$\textit{chain of thought}, output$\rangle$.
This presents us with the unique opportunity to observe the LLM's verbalized reasoning process in natural language when the CoT is visible \cite{korbak2025chain, baker2025monitoring}, allowing us to study in real-time how reasoning unfolds at inference, in response to a specific input.

This observable interface lets us investigate LLM outputs in several ways, wherein verbalized CoTs can be evaluated as their explanations through properties such as faithfulness, robustness, and utility, effectively serving as evidence for the model's reasoning and behavior \citep{jie2024interpretable,korbak2025chain}.
For example, approaches focused on reasoning correctness include process-supervision methods and step-level verifiers, which assess the semantic validity or consistency of individual intermediate steps, either to identify errors or to shape the reasoning produced during training and inference \citep{uesato2022solving,lightman2024let,zheng2024processbench,he2025can}.
However, these approaches rely on the verbalized trace as an informative account of the reasoning process, even though CoTs need not faithfully reflect the computations underlying the model's answer.

Such unfaithfulness can manifest in several forms \citep{barez2025chain}. Models may omit factors that causally influence their predictions or produce plausible post-hoc rationales for conclusions reached through other means \citep{turpin2023language}.
Intervention-based studies further show that the extent to which models rely on their stated reasoning varies across tasks and model scales \citep{lanham2023measuring}. Furthermore, applying optimization pressure directly to intermediate reasoning may improve its apparent correctness or usefulness while reducing what the verbalized CoT reveals about the computations producing the answer \citep{baker2025monitoring,korbak2025chain}.
We therefore cannot straightforwardly treat CoT as a literal transcript or mechanistic explanation of the model's internal reasoning.

Nevertheless, the absence of guaranteed faithfulness does not preclude verbalized CoTs from containing useful behavioral evidence about reasoning capabilities.
Indeed, recent work has begun to examine reasoning beyond the correctness of isolated statements;
E.g., causal analyses identify individual reasoning steps that disproportionately influence distant downstream reasoning and show that understanding such effects may require resampling alternative continuations rather than analyzing a single realized CoT \citep{bogdan2025thought,macar2025thought}.
Other work demonstrates that early reasoning errors can shape the remainder of a trajectory \citep{liao2025lost}, analyzes how the influence and faithfulness of CoT evolve over the course of reasoning \citep{lewis2025analysing}, and finds that the trajectory of answer uncertainty can predict final reasoning reliability \citep{zhao2026entropy}. Together, these works show that the location, progression, and broader organization of reasoning can provide information not captured by evaluating individual statements independently.

Complementary approaches study trajectories through either internal representations or structured abstractions of visible CoT and have established reasoning dynamics as an active area of investigation. For example, hidden-state analyses find that correct and incorrect reasoning trajectories diverge in representation space and use this divergence for mid-reasoning prediction and steering \citep{sun2026llm}.
Analyses of visible reasoning traces have annotated transitions between cognitive episodes \citep{li2025understanding}, induced recurring reasoning operators that support correctness prediction before a trace is complete \citep{lee2026reasonops}, and characterized early failure transitions that can guide uncertainty-based intervention \citep{zhu2026dissecting}. However, how task-dependent capability failures manifest through the sequencing, recurrence, and timing of visible reasoning functions remains poorly characterized, and whether these signatures can subsequently motivate procedure-specific correction remains unclear.

{In this work, we therefore ask, without assuming that verbalized reasoning faithfully reflects the model's internal computations:} {i) How do task-dependent capability failures manifest in the dynamics of visible CoT,
ii) Can these signatures be identified before answer emission,
and iii) Can they guide procedure-specific correction?}

To answer these questions, we study five different LLM configurations on Boolean satisfiability problems \citep{2025arXiv250514615W}
with systematically variable complexity. This setting provides externally established ground truth, with some instances admitting compact, solver-checkable solutions and others requiring more open-ended, proof-oriented reasoning that better approximates complex real-world demands, while enabling controlled comparisons that can be adapted to each model's reasoning capabilities.

To study the generated CoTs, we segment and tag sentences in each trace according to their reasoning function \citep{bogdan2025thought,venhoff2025understanding}, and analyze the role densities, transitions, cycles, entropy, and finalization timing.
Through this framework, we find that capability failures manifest as structured, task-dependent shifts in visible CoT dynamics; these shifts can provide warning that the reasoning process is likely to culminate in an incorrect answer and, when they reveal a mismatch between the procedure used and the demands of the task, motivate targeted correction.
Specifically, our contributions include:
\begin{itemize}
\item \textbf{A capability-matched framework for studying reasoning failure.}
We identify model-specific capability-frontier regimes in which each model produces both successful and failed traces on comparable problems. This provides controlled settings for analyzing capability breakdown without conflating failure with either near-perfect task mastery or broad inability.

\item \textbf{A recurring SAT failure mode characterized by premature verification collapse.}
On matched satisfiable (SAT) instances, incorrect traces frequently enter clause checking earlier, revisit similar reasoning operations more often, and finalize sooner than successful traces despite comparable clause coverage. We refer to this model-dependent trajectory pattern as \emph{premature verification collapse}, showing that failure can be reflected in how reasoning operations are organized rather than simply whether the model attends to the relevant clauses.

\item \textbf{Early-warning signals from visible reasoning dynamics.}
By tracking reasoning-dynamics features as the CoT unfolds and evaluating them on held-out problems, we show that many incorrect traces can be identified before final-answer emission.

\item \textbf{A task-specific reasoning-mode mismatch and diagnosis-informed correction.}
On unsatisfiable (UNSAT) problems, models often organize their reasoning around proposing and checking candidate assignments by incorrectly assuming and concluding the problem is SAT, rather than systematically constructing cases and deriving contradictions that allow it to be proven as UNSAT.
Motivated by this mismatch between the procedure used and the proof demanded by the task, we introduce a targeted prompt that elicits case splitting, consequence propagation, and contradiction search. This intervention raises Llama3-70B accuracy from $13.3\%$ to $85.0\%$, correcting $84.6\%$ of the originally incorrect SAT conclusions.

\end{itemize}

Together, these results show that capability failures can indeed manifest not only in final outcomes or isolated incorrect steps, but also as distributed, task-dependent changes in the sequencing, repetition, and timing of visible reasoning. Even when CoTs cannot be interpreted as a faithful transcript of internal computation, their dynamics can provide useful behavioral evidence about how reasoning is succeeding or failing, offering a complementary signal for monitoring and correcting failures of reasoning LLMs.

\section{Methodology}
\label{sec:method}

In this section, we describe how we construct the task setup, characterize reasoning dynamics from visible CoT, and evaluate whether the resulting signatures support diagnosis, early warning, and correction.

\begin{figure*}[t]
    \centering
    \includegraphics[height=0.52\textheight,keepaspectratio]{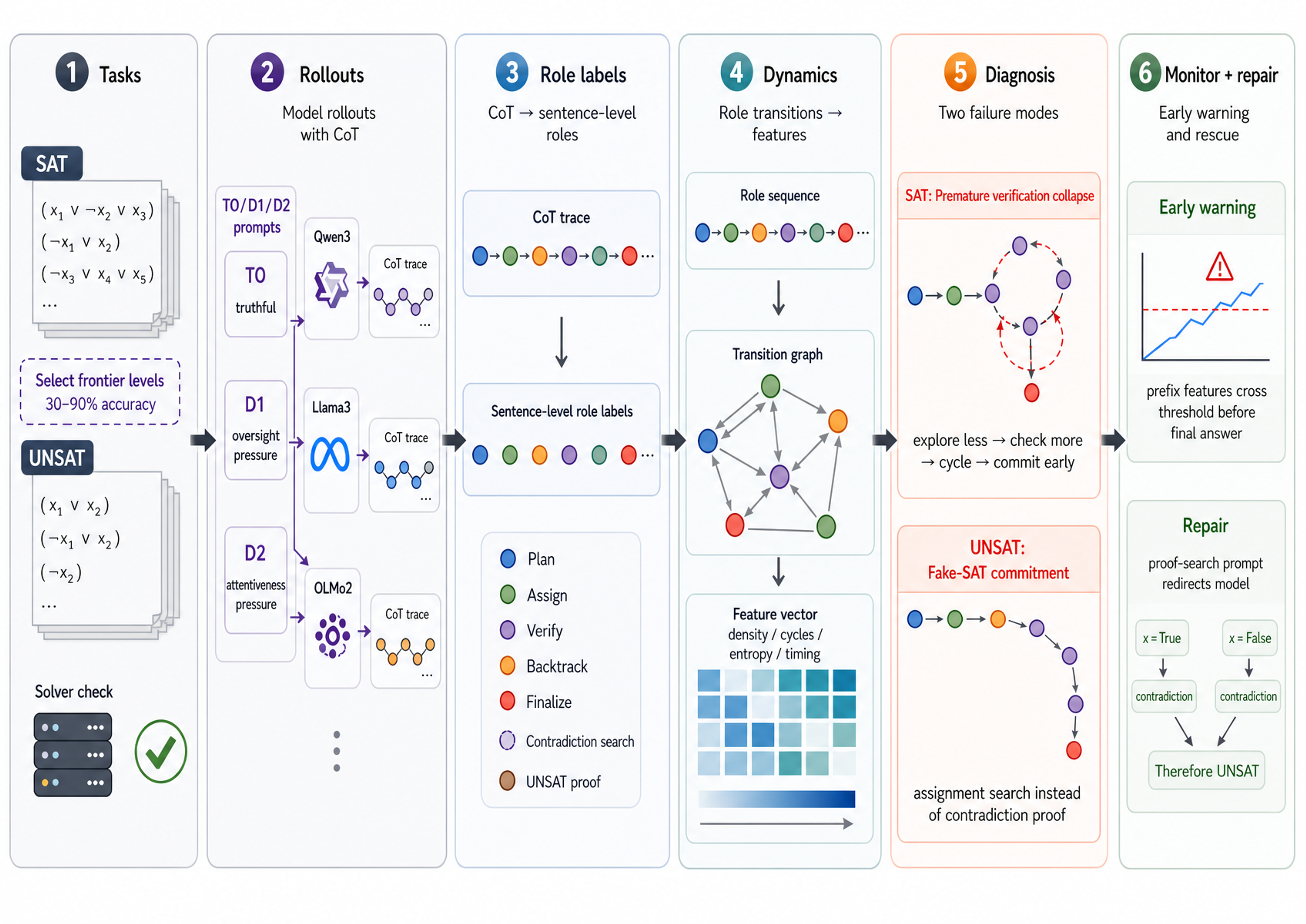}
    \caption{Analysis pipeline. We generate solver-certified SAT and UNSAT tasks, select model-specific mixed-success frontier levels, collect CoT rollouts under three prompt framings, map sentences to coarse reasoning roles, summarize role dynamics, diagnose SAT and UNSAT failure modes, evaluate prefix-based warning, and test a post-hoc proof-search repair.}
    \label{fig:pipeline_overview}
\end{figure*}

\subsection{Tasks and experimental setup}

We study reasoning on Boolean satisfiability problems, in which each task is a formula in conjunctive normal form and the model must decide whether any assignment of truth values to variables satisfies every clause simultaneously.
We adopt this setting for three reasons:
i) Every instance carries a label that an external solver establishes independently of anything the model writes, so correctness can be assessed without treating the visible trace as evidence about itself,
ii) Instance complexity can be varied systematically, which allows us to place each model at a difficulty where it both succeeds and fails rather than comparing models at a fixed level that is trivial for one and unreachable for another,
iii) Satisfiable and unsatisfiable instances impose asymmetric demands within a single formalism: a satisfiable (SAT) instance is settled by exhibiting one assignment that a solver checks directly, whereas an unsatisfiable (UNSAT) instance requires ruling out the entire assignment space and therefore demands case construction and contradiction rather than candidate proposal.
This asymmetry gives us two distinct reasoning procedures to compare while holding the task format, prompts, and verification machinery fixed.
Correctness is accordingly determined; a SAT response counts as correct only when the model reports SAT and supplies an assignment that the solver verifies against every clause, while for UNSAT instances the solver provides the ground-truth label and we evaluate the reported label without claiming that the accompanying argument constitutes a formal proof.

At inference, we assign each rollout one of four outcomes: \texttt{correct\_complete}, \texttt{wrong\_complete}, \texttt{truncated\_or\_looping}, or \texttt{format\_failure}. The first two require a complete, parseable answer and differ only in solver-verified correctness, whereas truncated, looping, and unparseable responses are retained separately rather than merged with complete but incorrect reasoning, since they reflect failures of generation rather than of reasoning. The UNSAT analysis focuses on \texttt{wrong\_sat\_complete}, a complete SAT claim on a solver-verified unsatisfiable formula, in which the model reaches an assignment-based conclusion on a task that instead demands an impossibility argument.

We evaluate Qwen3-8B, Qwen3-14B, Llama3-8B, Llama3-70B, and OLMo2-13B, sampling each fixed formula under three prompt conditions: a baseline solving prompt (T0), one stating that an overseer is evaluating whether the reasoning appears careful and reliable (D1), and one emphasizing attentiveness through clause tracking, assumption checking, and explicit work (D2). Since D1 and D2 hold both the formula and the required answer fixed, we treat them as controlled framing perturbations rather than manipulations assumed to induce deception or a clean hidden-belief/report mismatch. All  task and intervention prompts, outcome definitions, and additional
experimental details are provided in supplementary material (Sections~A-B).





\subsection{Capability frontiers and capability matching}

\begin{figure}[t]
    \centering
    \includegraphics[width=\linewidth]{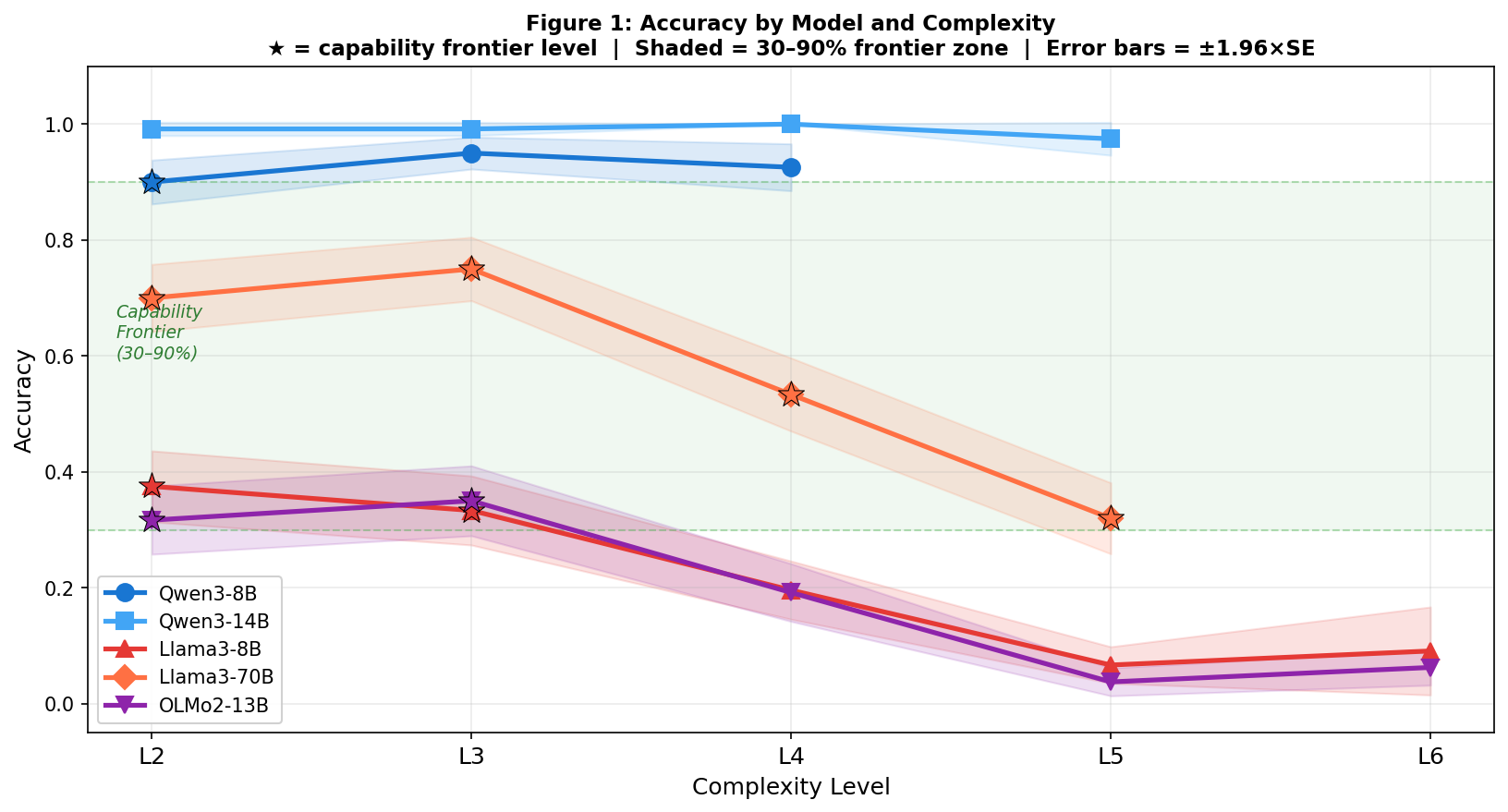}
    \caption{Accuracy across SAT complexity levels. The shaded 30--90\% band marks the mixed-success frontier used for controlled comparisons.}
    \label{fig:capability_frontier}
\end{figure}

Because a level that is nearly trivial for one model (e.g., Qwen3-14B) may already lie beyond the reliable range of (e.g., Llama3-8B), comparing models at a single fixed difficulty would yield almost no failures for one and almost no successes for the other.
We therefore select difficulty separately for each model, following the item-response intuition that examples are most informative when success is neither certain nor impossible \citep{lord1980applications,baker2001basics}.

Letting $\widehat{p}_{m,\ell}$ denote the empirical accuracy of model $m$ at complexity level $\ell$, we take the model--level pair to lie on the capability frontier when $0.30 \leq \widehat{p}_{m,\ell} \leq 0.90$, where the lower bound requires a nontrivial number of successful traces and the upper bound leaves enough failures for analysis.

This band is an ad hoc rule for the present study rather than a universal capability boundary.
Figure~\ref{fig:capability_frontier} reports all tested levels and shows that a single complexity level places models in different regimes:
Qwen3-14B remains above the frontier throughout the sweep, Llama3-8B and OLMo2-13B fall below it at harder levels, and Llama3-70B stays within it from L2 through L5.
Of 127 matched problem--condition pairs, 91 are divergent and concentrate at intermediate levels (25 at L2, 27 at L3, 29 at L4, 10 at L5), supplying controlled comparisons near each model's capability boundary rather than contrasts between trivial successes and broad failures.

\subsection{Reasoning dynamics and analysis protocols}

We split each visible CoT into sentence-level units following the literature \citep{bogdan2025thought,venhoff2025understanding}, and assign each sentence a coarse functional role using a rule-based regex classifier.
For traces on SAT instances the roles are \emph{planning}, \emph{assignment}, \emph{verification}, \emph{backtracking}, \emph{finalization}, and \emph{other};
UNSAT traces additionally use proof-oriented roles for contradiction search, explicit UNSAT proof construction, SAT commitment, and clause-level checking.

For a role sequence $r_1,\ldots,r_T$ we count transitions $r_t\rightarrow r_{t+1}$ and form the row-normalized matrix $P_{ij}=\Pr(r_{t+1}=j\mid r_t=i)$, from which we compute role densities, cycle rate (returns to recent role patterns), weighted self-transition (persistence in the same role), transition entropy (diversity in next-role choices), and finalization timing (normalized position where answer-emission begins), together with contradiction-search and UNSAT-proof densities on unsatisfiable instances.

We then compare these features across divergent cases, in which the instance and prompt condition are held fixed while one model gives a wrong complete answer and another a correct one.
Our principal comparison is Llama3-8B wrong versus Qwen3-14B correct, which we supplement with OLMo2-13B wrong versus Qwen3-14B correct and, to reduce family-style confounding, Llama3-8B wrong versus Llama3-70B correct. Within-model comparisons provide a tighter control for model-specific response style (Supplementary Section~E.1). We observe that wrong traces consistently show higher backtracking, while the length, verification, cycling, and entropy effects vary across models.
Paired Wilcoxon signed-rank tests are computed over matched problem--condition pairs with paired bootstrap confidence intervals; the cross-family comparisons identify the clearest pattern, while the within-family comparison tests which parts persist when response style is more closely matched.

To test whether these signatures appear before the answer, we recompute the same features over increasing fractions of each trace and flag a trace once its score crosses a threshold. Because typical feature values differ across models and difficulty levels, both the rescaling and the thresholds are derived from each model's own correct traces under the baseline prompt. We split problems in half, using one half to fix these choices and the other only for evaluation, so no formula contributes to both; lead time is then the number of sentences between the first flag and the final answer.

\subsection{UNSAT comparisons and intervention design}

Since UNSAT instances demand an impossibility argument rather than a candidate assignment, we compare traces that wrongly conclude SAT against correct UNSAT traces to test whether the former search less for contradictions, build fewer explicit proofs, and instead check candidates, cycle, and commit to SAT earlier.
Any such difference could arise trivially, since a trace that concludes SAT contains language announcing the same.
We therefore repeat the comparison twice, once with explicit final-answer and commitment sentences removed and once with all answer-related features excluded, so that any remaining separation reflects how the reasoning is organized rather than what it reports.

If the comparison does reveal a mismatch between the procedure a model follows and the one the task demands, the natural next question is whether that procedure can be changed.
We therefore rerun the Llama3-70B wrong-SAT cases under two prompts;
A generic retry asks for another careful attempt without naming a strategy, while a targeted proof-search prompt asks the model to split into cases, propagate forced consequences, derive contradictions, and claim SAT only after checking a complete assignment.
Comparing the two separates whether these failures are simply unstable under resampling vs. whether they respond specifically to being directed toward the procedure that the diagnosis otherwise identifies as missing.

\section{Results}

\subsection{SAT failures show a recurring collapse in reasoning dynamics}

\begin{figure}[t]
    \centering
    \includegraphics[width=\linewidth]{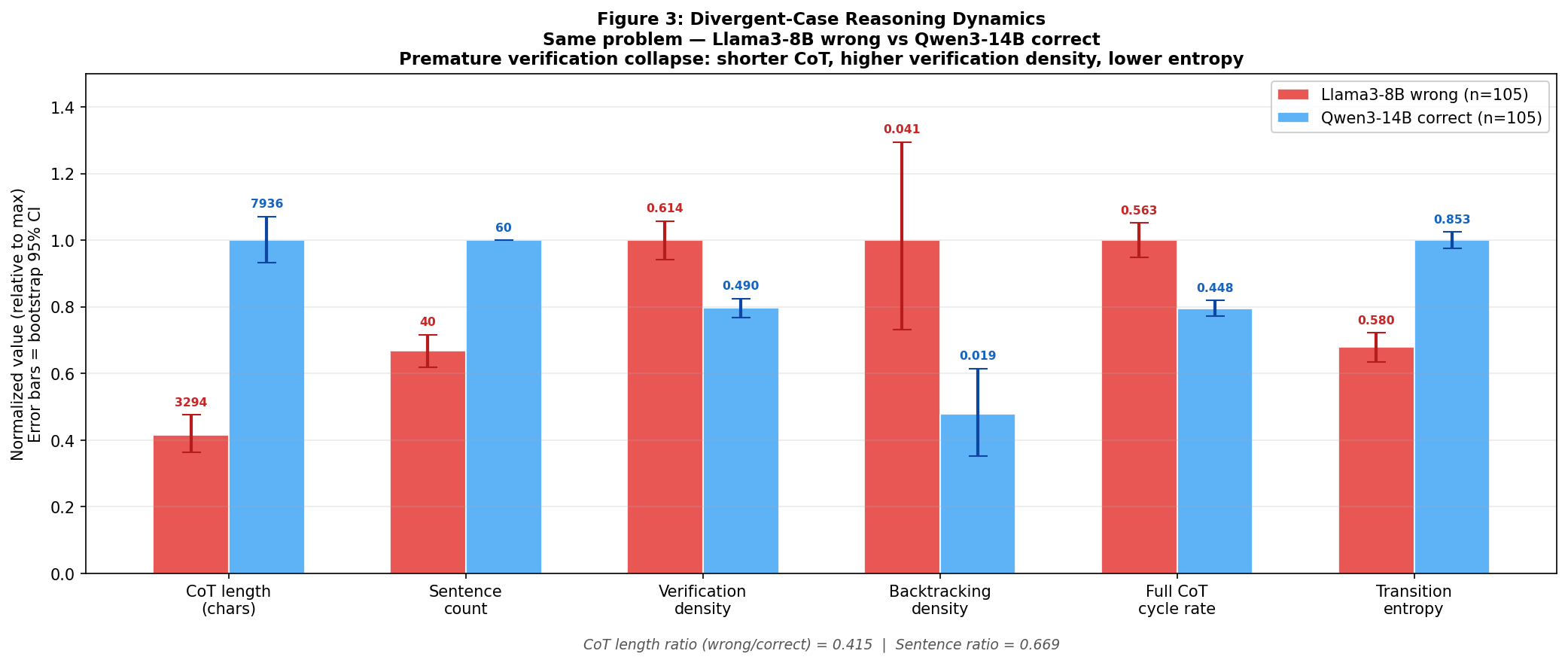}
    \caption{Matched SAT dynamics for Llama3-8B wrong versus Qwen3-14B correct traces on the same problem and prompt condition. Failing traces are shorter, more verification-heavy and cyclic, lower-entropy, and earlier-finalizing.}
    \label{fig:divergent_dynamics}
\end{figure}



Across 105 matched Llama3-8B-wrong/Qwen3-14B-correct pairs, wrong traces are roughly half as long, more verification-heavy and more cyclic, and markedly lower in transition entropy (Table~\ref{tab:matched_divergent_pairs}), beginning finalization at 77.1\% of the trace rather than 89.8\%. 

Figure~\ref{fig:divergent_dynamics} shows the differences in trace length, sentence count, verification and backtracking density, cycle rate, and transition entropy. Clause coverage, however, does not differ ($p=0.206$), so the failing traces do not simply ignore more of the formula; they mention much of the same clauses while organizing their reasoning around narrower, repeated checking and earlier commitment.
We call this pattern \emph{premature verification collapse}: the failing trace settles into a repetitive verification mode before sustaining enough exploration or revision to support the answer it commits to.

\begin{table*}[t]
\centering
\small
\setlength{\tabcolsep}{4pt}
\caption{Matched divergent-case comparisons. Each row holds the SAT problem and prompt condition fixed while the first model is wrong and the second is correct. Cross-family comparisons show the clearest collapse; the within-family Llama comparison is more mixed.}
\label{tab:matched_divergent_pairs}
\begin{tabular}{llrrrrc}
\toprule
Pair & Metric & Wrong $\mu$ & Correct $\mu$ & $\Delta$ & $p$ & Sig. \\
\midrule
\multicolumn{7}{l}{\textbf{Llama3-8B wrong vs. Qwen3-14B correct (cross-family)}} \\
& CoT length (chars)   & 3293.60 & 7936.14 & -4642.54 & $<.001$ & *** \\
& Sentence count       & 40.11 & 60.00 & -19.89 & $<.001$ & *** \\
& Verification density & 0.61 & 0.49 & +0.12 & $<.001$ & *** \\
& Backtracking density & 0.04 & 0.02 & +0.02 & .0013 & ** \\
& Full-CoT cycle rate  & 0.56 & 0.45 & +0.11 & $<.001$ & *** \\
& Transition entropy   & 0.58 & 0.85 & -0.27 & $<.001$ & *** \\
\midrule
\multicolumn{7}{l}{\textbf{Llama3-8B wrong vs. Llama3-70B correct (within-family)}} \\
& CoT length (chars)   & 3554.67 & 2284.23 & +1270.44 & 1.0000 & ns \\
& Sentence count       & 41.90 & 32.45 & +9.45 & 1.0000 & ns \\
& Verification density & 0.61 & 0.64 & -0.03 & .0427 & * \\
& Backtracking density & 0.04 & 0.01 & +0.03 & $<.001$ & *** \\
& Full-CoT cycle rate  & 0.56 & 0.56 & +0.01 & .0471 & * \\
& Transition entropy   & 0.57 & 0.60 & -0.03 & .0133 & * \\
\midrule
\multicolumn{7}{l}{\textbf{OLMo2-13B wrong vs. Qwen3-14B correct (cross-family)}} \\
& CoT length (chars)   & 2059.32 & 7900.68 & -5841.37 & $<.001$ & *** \\
& Sentence count       & 26.84 & 60.00 & -33.16 & $<.001$ & *** \\
& Verification density & 0.74 & 0.49 & +0.25 & $<.001$ & *** \\
& Backtracking density & 0.03 & 0.02 & +0.01 & .0825 & $\dagger$ \\
& Full-CoT cycle rate  & 0.63 & 0.45 & +0.18 & $<.001$ & *** \\
& Transition entropy   & 0.44 & 0.85 & -0.42 & $<.001$ & *** \\
\bottomrule
\end{tabular}

\footnotesize{*** $p<.001$, ** $p<.01$, * $p<.05$, $\dagger$ $p<.10$; ns = not significant. One-sided $p=1.0000$ indicates an effect opposite to the tested direction.}
\end{table*}


Table~\ref{tab:matched_divergent_pairs} qualifies how far this pattern generalizes. Increased cycling and reduced transition entropy hold in both the cross-family and within-family comparisons, whereas the shorter, more verification-heavy profile appears only cross-family, most strongly for OLMo2-13B against Qwen3-14B. Repetition and reduced variability are therefore the robust signature, while the accompanying length and verification effects depend on model family and response style.

\subsection{Reasoning dynamics provide model-dependent early warnings}

\begin{figure}[t]
    \centering
    \includegraphics[width=0.95\linewidth]{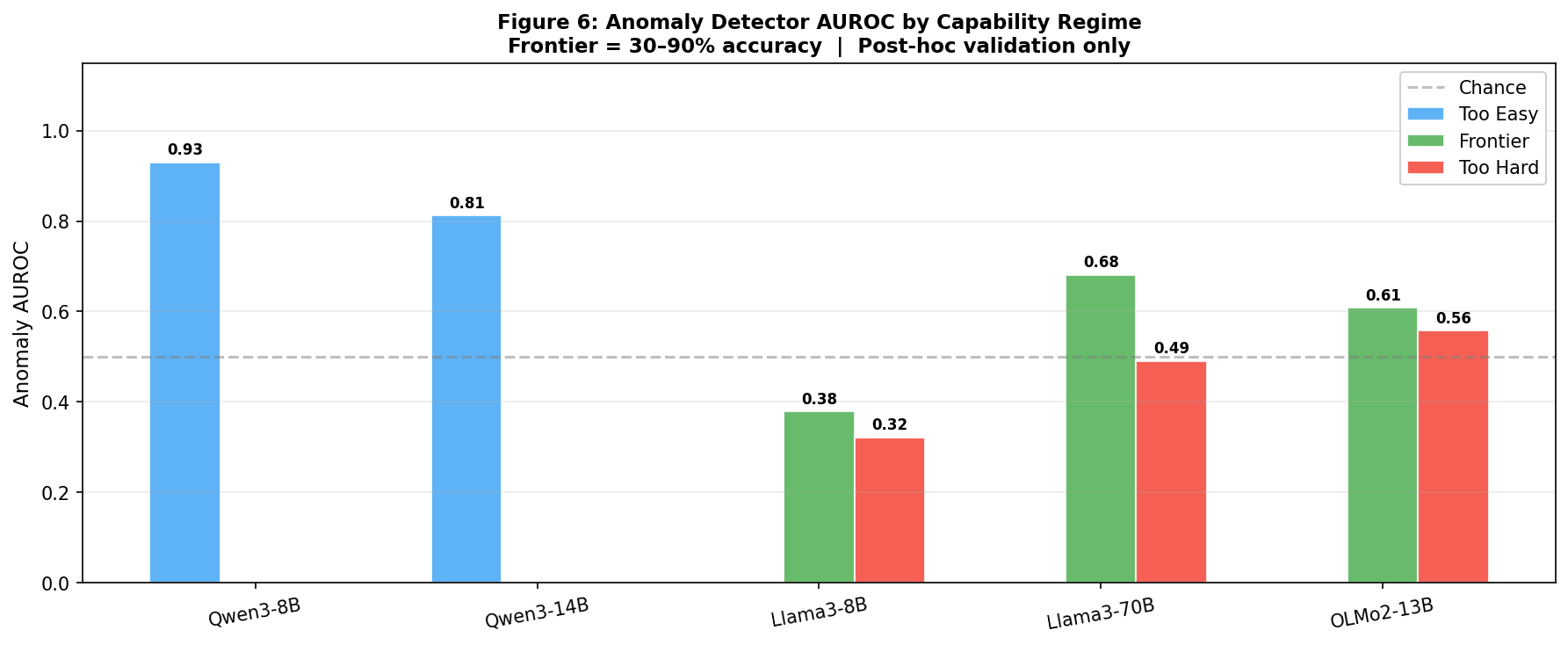}
    \caption{Detector performance by capability regime. AUROC is most interpretable in mixed-success frontier levels, where both correct and wrong complete traces are available.}
    \label{fig:auroc_by_regime}
\end{figure}

\begin{figure*}[t]
    \centering
    \begin{minipage}[t]{0.5\textwidth}
        \centering
        \includegraphics[width=\linewidth]{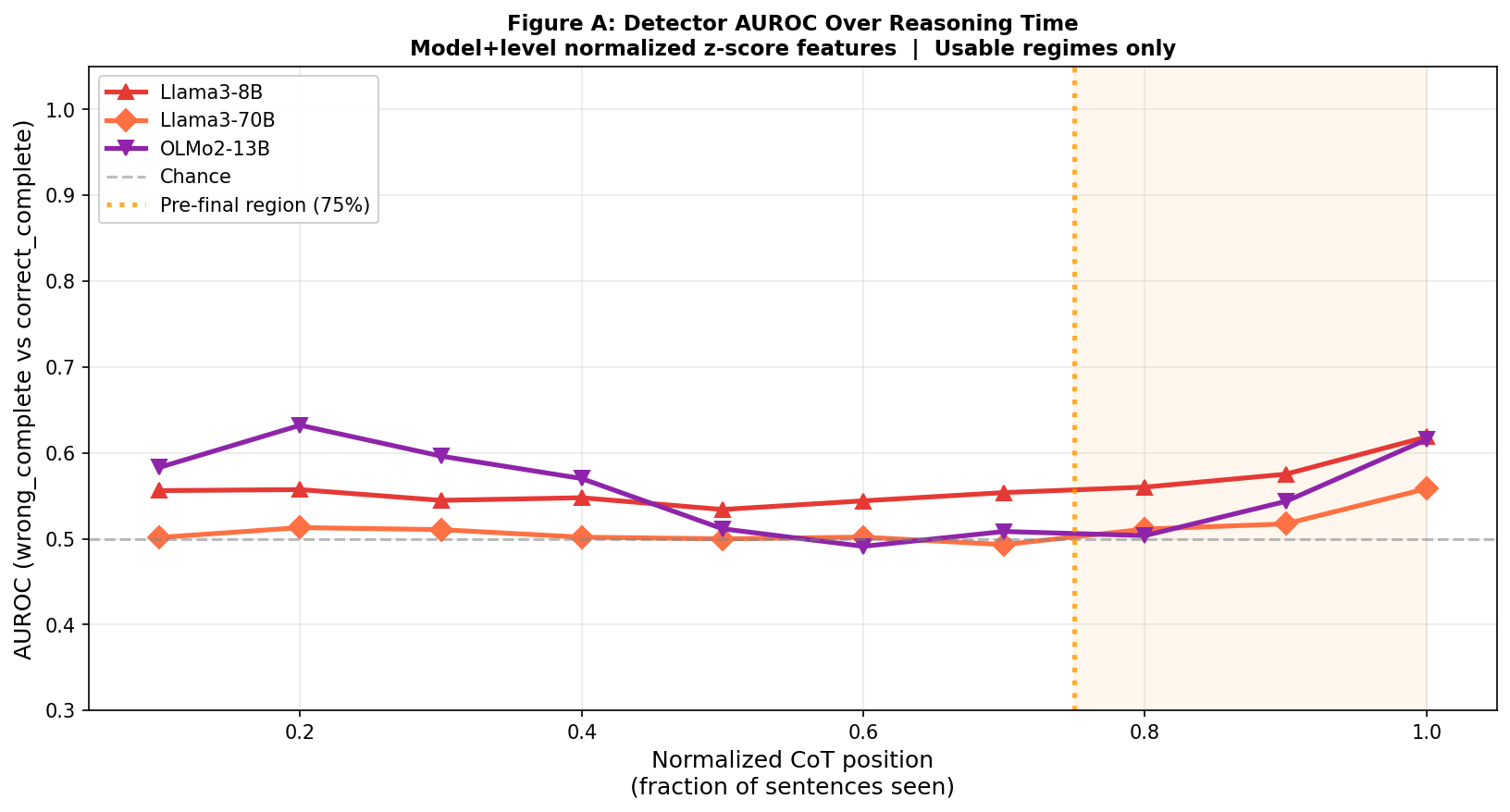}
    \end{minipage}\hfill
    \begin{minipage}[t]{\textwidth}
        \centering
        \includegraphics[width=0.65\linewidth]{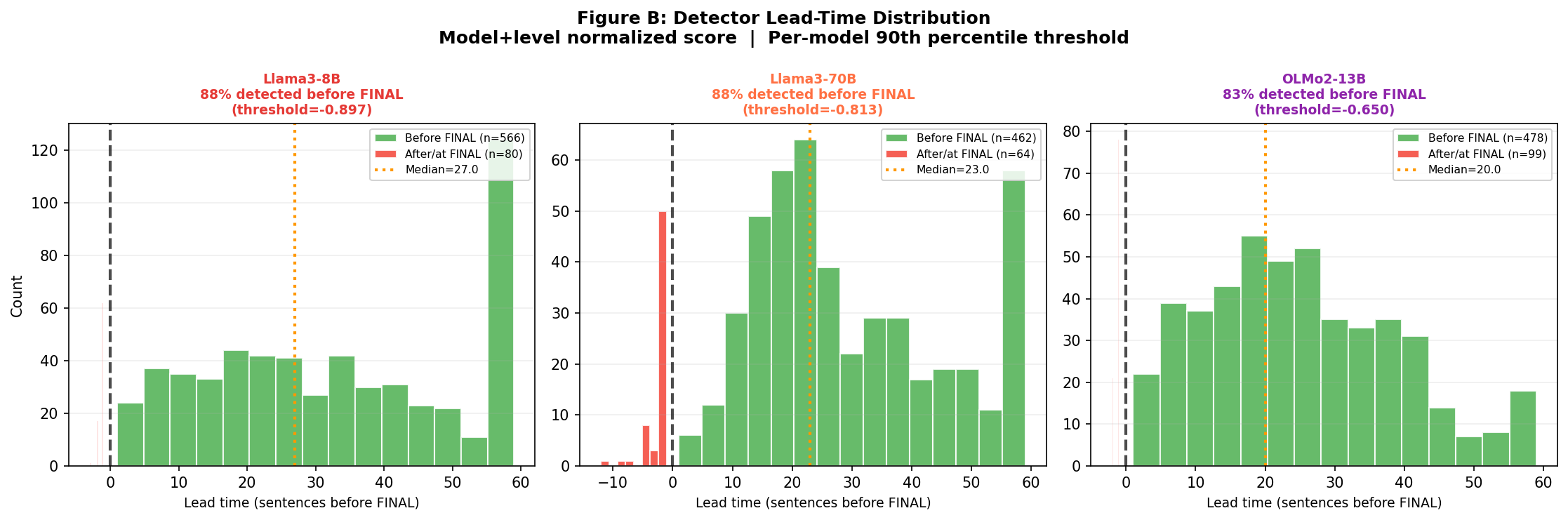}
    \end{minipage}
    \caption{Held-out early warning. Left: detector AUROC over normalized CoT position. Right: fraction of wrong-complete traces flagged before the final answer and lead time. The signal is useful for some models but is not universal.}
    \label{fig:detector_combined}
\end{figure*}

We find that whether failures can be flagged in advance as the CoT unfolds depends first on where a model sits relative to its capability frontier (Figure~\ref{fig:auroc_by_regime}).
Nearly saturated settings yield too few complete errors for stable evaluation, whereas in broadly failing settings errors need not appear anomalous relative to the model's usual behavior.
Hence, the frontier is the regime in which a calibrated score has a meaningful reference distribution.

Scoring partial traces at increasing fractions of their length, we find the score rises above chance near the end of the trace for Llama3-8B and OLMo2-13B, while Llama3-70B remains weaker (Figure~\ref{fig:detector_combined}, left).
At the threshold set on the calibration half, 80.2\% of Llama3-8B wrong-complete traces are flagged before the final answer with a median lead time of 23 sentences, and 78.2\% of OLMo2-13B traces with 17.5 sentences, whereas Llama3-70B reaches only 35.7\% with zero median lead time (Figure~\ref{fig:detector_combined}, right).
Visible-dynamics features can therefore provide advance warning in some model and difficulty regimes, but the present score is not a general failure detector, and the higher coverage figures should be read together with the per-model calibration and held-out split rather than as a universal operating point.

\subsection{UNSAT failures substitute assignment search for contradiction proof}

\begin{figure*}[t]
    \centering
    \begin{minipage}[t]{0.5\textwidth}
        \centering
        \includegraphics[width=\linewidth]{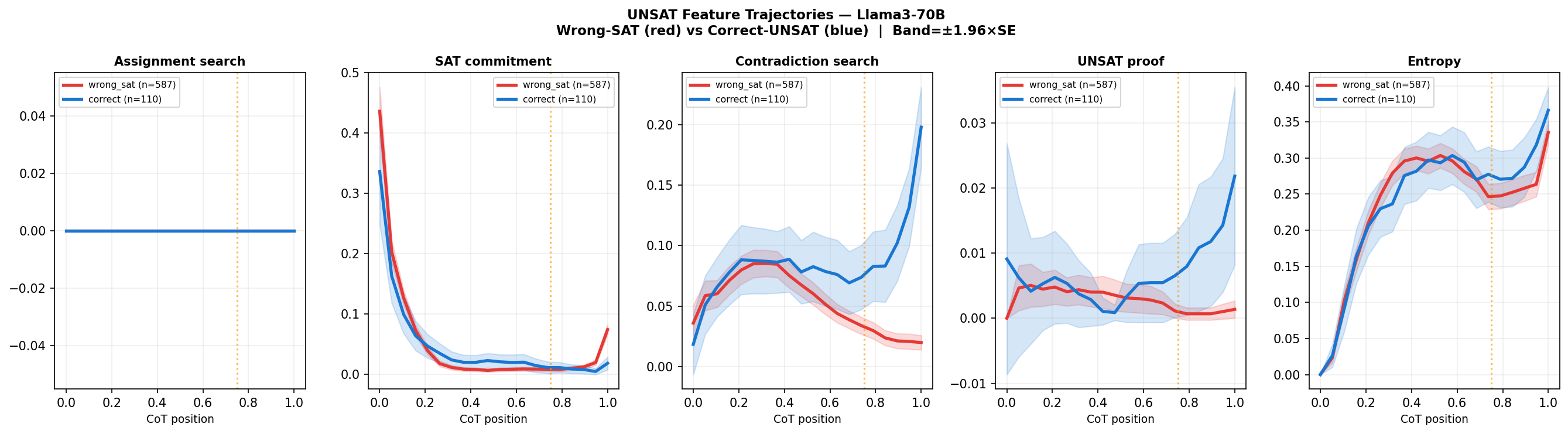}
    \end{minipage}\hfill
    \begin{minipage}[t]{0.5\textwidth}
        \centering
        \includegraphics[width=\linewidth]{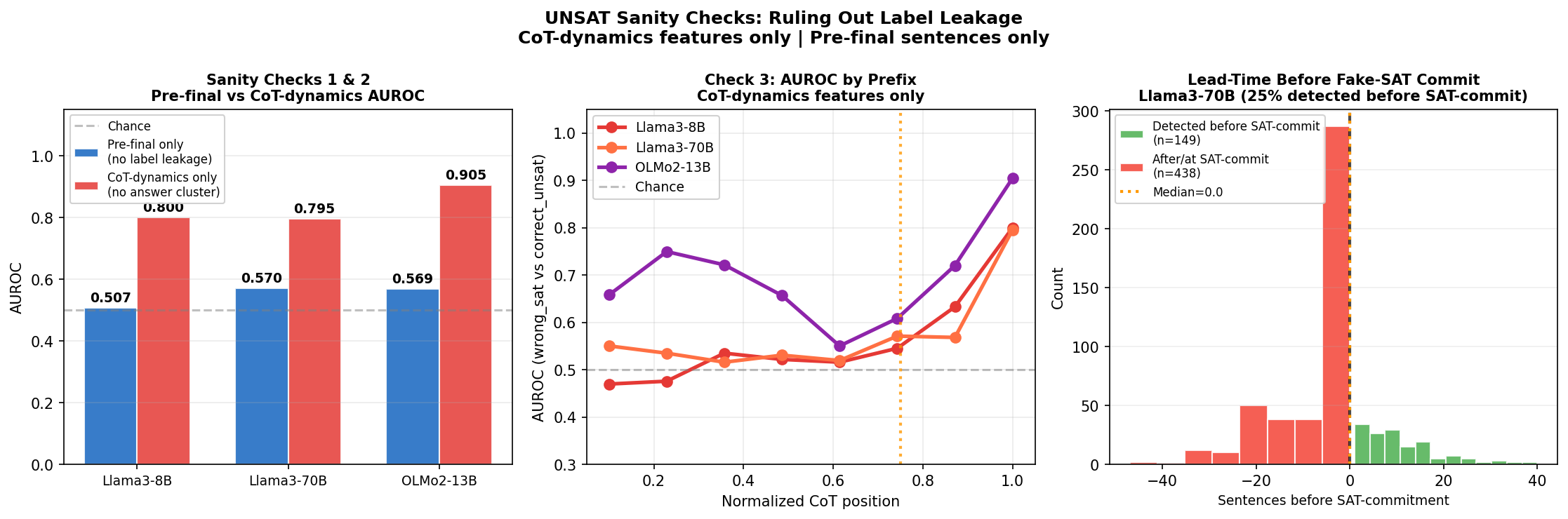}
    \end{minipage}
    \caption{UNSAT diagnosis. Left: wrong-SAT traces show less contradiction search and UNSAT-proof behavior and more assignment-verification behavior than correct-UNSAT traces. Right: dynamics-only features remain predictive after excluding final-label features, while pre-final-only discrimination is modest.}
    \label{fig:unsat_combined}
\end{figure*}




Relative to SAT instances, UNSAT failures nearly always take the same form: rather than running out of time or declining to answer, Llama and OLMo models assert that the formula is SAT.
Llama3-70B is most affected, its frontier accuracy falling from 0.526 on SAT to 0.159 on UNSAT, with false SAT claims accounting for 0.800 of its completions. Because this occurs under every prompt condition, and becomes more frequent for Llama3-70B under the D1 and D2 framings, it is not an artifact of one baseline prompt.

The role trajectories locate the mismatch (Figure~\ref{fig:unsat_combined}, left): wrong-SAT traces contain less contradiction search or explicit UNSAT-proof construction, and instead more assignment verification and cycling, as though the model were approaching an impossibility proof by searching for a single candidate assignment.
This reading does not follow from the final label alone;
After removing answer-related features, AUROC remains 0.800 for Llama3-8B, 0.795 for Llama3-70B, and 0.905 for OLMo2-13B, so the separation reflects how the reasoning is organized rather than what it reports.
Discrimination from the sentences preceding the answer alone is nonetheless modest (AUROC 0.507, 0.570, and 0.569): the separation emerges only once the full trace is available, so these features explain UNSAT failures after the fact but cannot flag them while the model is still reasoning.

\subsection{Prompting for proof search rescuers most UNSAT failures}

\begin{table}[t]
\centering
\small
\setlength{\tabcolsep}{4pt}
\caption{Llama3-70B post-hoc intervention on the same UNSAT cases. Correction rate is computed over the 52 original wrong-SAT completions.}
\label{tab:intervention}
\begin{tabular}{lrr}
\toprule
Condition & Accuracy & Corrected wrong-SAT \\
\midrule
Original & 13.3\% & -- \\
Generic retry & 10.0\% & 6/52 (11.5\%) \\
Proof-search prompt & \textbf{85.0\%} & \textbf{44/52 (84.6\%)} \\
\bottomrule
\end{tabular}
\end{table}



If the wrong-SAT completions reflect the wrong procedure rather than the wrong answer, directing the model toward case splitting and contradiction search should correct them.
Table~\ref{tab:intervention} shows that a generic retry does not, whereas the targeted proof-search prompt raises accuracy from 13.3\% to 85.0\%, correcting 44 of 52 original wrong-SAT completions, a paired improvement of 73.1 percentage points (bootstrap 95\% CI: 61.5--84.6) that is significant under McNemar's test. The visible trace shifts accordingly, with SAT-commitment density falling from 0.0263 to 0.0105 and contradiction-search density rising from 0.0658 to 0.1097, which links the gain to the diagnosed procedure rather than to a different final label alone.

This does not show that the model internally knew the correct answer, nor that repair can be triggered without ground truth; it shows that many failures are sensitive to the reasoning procedure the prompt elicits.

\section{Discussion and Limitations}

Given that our analysis reads the organization of emitted text alone, the interpretation it supports is a behavioral one: visible CoT carries useful information about how a model's reasoning is arranged, irrespective of the exact relation and faithfulness that arrangement bears to the computation producing the answer.
In the same vein, the SAT and UNSAT results also indicate that reasoning failure is not one homogeneous state, since failure on SAT instances involves a gradual narrowing into repetitive verification whereas failure on UNSAT instances reflects selection of the wrong procedure altogether.
Correction therefore has to be procedure-specific;
a generic request to try again reproduces the same mode, whereas an instruction aimed at the diagnosed procedure redirects the trace toward the operation it is missing.

The results these readings rest on, however, carry several qualifications. The strongest SAT comparison is cross-family, and since models differ systematically in verbosity and response style, part of the length and verification gap may reflect those differences rather than failure as such. The within-family analysis preserves the repetition and entropy effects but not the length and verification effects, so premature verification collapse is better understood as a recurring pattern than a universal signature.

Reasoning roles are likewise abstractions of visible text rather than observations of what a model is doing, and every dynamics feature we report inherits whatever error the tagger introduces; agreement statistics bound this but do not eliminate it. Early-warning performance also varies sharply across models, and the coverage figures we report fix a threshold without characterizing what it costs. A deployment-oriented evaluation would need false-positive rates, precision, coverage, and threshold sensitivity, together with simple baselines such as length, finalization timing, or role densities without transitions, to establish that the transition structure contributes beyond what these cheaper signals already provide.

The intervention is post hoc and oracle-assisted, since solver labels determine which failures are rerun, so what we establish is the correctability of a diagnosed failure mode rather than an end-to-end monitor-and-repair system. The generic retry is also unmatched to the proof-search prompt in length and detail, leaving open how much of the gain follows from UNSAT-specific guidance and how much from a longer, more explicit instruction. Joining the early-warning score to the intervention would address the first of these, testing whether repair can be triggered without ground-truth labels.

Finally, while Boolean satisfiability supplies exact labels, adjustable difficulty, and interpretable procedural demands, generality to open-ended domains remains to be shown. Extending the analysis requires tasks that combine externally checkable outcomes with a meaningful distinction between alternative reasoning procedures, a combination that is uncommon outside formal settings. Within these limits, visible trace dynamics can complement semantic step checking by revealing failures distributed across the trajectory.

\section{Conclusion}

Studying how visible reasoning is organized rather than whether each of its statements is correct, we find that failures need not be localized to a single incorrect sentence: on SAT instances failed traces become more repetitive and less varied and can collapse prematurely into verification and finalization, whereas on UNSAT instances models may select the wrong procedure altogether, verifying candidate assignments instead of constructing contradictions. These dynamics sometimes provide advance warning and, more consequentially, identify a concrete reasoning procedure whose elicitation rescues most observed wrong-SAT failures. Visible CoT therefore remains useful as a behavioral monitoring signal even where it cannot be treated as a faithful transcript of internal reasoning.

\clearpage
\bibliography{aaai2027}

@ARTICLE{2025arXiv250514615W,
       author = {{Wei}, Anjiang and {Wu}, Yuheng and {Wan}, Yingjia and {Suresh}, Tarun and {Tan}, Huanmi and {Zhou}, Zhanke and {Koyejo}, Sanmi and {Wang}, Ke and {Aiken}, Alex},
        title = "{SATBench: Benchmarking LLMs' Logical Reasoning via Automated Puzzle Generation from SAT Formulas}",
      journal = {arXiv e-prints},
         year = 2025,
        month = may,
          eid = {arXiv:2505.14615},
        pages = {arXiv:2505.14615},
          doi = {10.48550/arXiv.2505.14615},
archivePrefix = {arXiv},
       eprint = {2505.14615},
 primaryClass = {cs.AI},
       adsurl = {https://ui.adsabs.harvard.edu/abs/2025arXiv250514615W}
}

@book{lord1980applications,
  title={Applications of Item Response Theory to Practical Testing Problems},
  author={Lord, Frederic M.},
  year={1980},
  publisher={Lawrence Erlbaum Associates}
}

@book{baker2001basics,
  title={The Basics of Item Response Theory},
  author={Baker, Frank B.},
  year={2001},
  publisher={ERIC Clearinghouse on Assessment and Evaluation}
}

@article{wei2022chain,
  title={Chain-of-thought prompting elicits reasoning in large language models},
  author={Wei, Jason and Wang, Xuezhi and Schuurmans, Dale and Bosma, Maarten and Xia, Fei and Chi, Ed and Le, Quoc V and Zhou, Denny and others},
  journal={Advances in neural information processing systems},
  volume={35},
  pages={24824--24837},
  year={2022}
}

@article{jaech2024openai,
  title={Openai o1 system card},
  author={Jaech, Aaron and Kalai, Adam and Lerer, Adam and Richardson, Adam and El-Kishky, Ahmed and Low, Aiden and Helyar, Alec and Madry, Aleksander and Beutel, Alex and Carney, Alex and others},
  journal={arXiv preprint arXiv:2412.16720},
  year={2024}
}

@article{yang2025qwen3,
  title={Qwen3 technical report},
  author={Yang, An and Li, Anfeng and Yang, Baosong and Zhang, Beichen and Hui, Binyuan and Zheng, Bo and Yu, Bowen and Gao, Chang and Huang, Chengen and Lv, Chenxu and others},
  journal={arXiv preprint arXiv:2505.09388},
  year={2025}
}

@article{korbak2025chain,
  title={Chain of thought monitorability: A new and fragile opportunity for ai safety},
  author={Korbak, Tomek and Balesni, Mikita and Barnes, Elizabeth and Bengio, Yoshua and Benton, Joe and Bloom, Joseph and Chen, Mark and Cooney, Alan and Dafoe, Allan and Dragan, Anca and others},
  journal={arXiv preprint arXiv:2507.11473},
  year={2025}
}

@article{baker2025monitoring,
  title={Monitoring reasoning models for misbehavior and the risks of promoting obfuscation},
  author={Baker, Bowen and Huizinga, Joost and Gao, Leo and Dou, Zehao and Guan, Melody Y and Madry, Aleksander and Zaremba, Wojciech and Pachocki, Jakub and Farhi, David},
  journal={arXiv preprint arXiv:2503.11926},
  year={2025}
}

@article{uesato2022solving,
  title={Solving math word problems with process-and outcome-based feedback},
  author={Uesato, Jonathan and Kushman, Nate and Kumar, Ramana and Song, Francis and Siegel, Noah and Wang, Lisa and Creswell, Antonia and Irving, Geoffrey and Higgins, Irina},
  journal={arXiv preprint arXiv:2211.14275},
  year={2022}
}

@inproceedings{lightman2024let,
  title={Let's verify step by step},
  author={Lightman, Hunter and Kosaraju, Vineet and Burda, Yuri and Edwards, Harrison and Baker, Bowen and Lee, Teddy and Leike, Jan and Schulman, John and Sutskever, Ilya and Cobbe, Karl},
  booktitle={International Conference on Learning Representations},
  volume={2024},
  pages={39578--39601},
  year={2024}
}

@article{zheng2024processbench,
  title={Processbench: Identifying process errors in mathematical reasoning},
  author={Zheng, Chujie and Zhang, Zhenru and Zhang, Beichen and Lin, Runji and Lu, Keming and Yu, Bowen and Liu, Dayiheng and Zhou, Jingren and Lin, Junyang},
  journal={arXiv preprint arXiv:2412.06559},
  year={2024}
}

@inproceedings{he2025can,
  title={Can large language models detect errors in long chain-of-thought reasoning?},
  author={He, Yancheng and Li, Shilong and Liu, Jiaheng and Wang, Weixun and Bu, Xingyuan and Zhang, Ge and Peng, Zy and Zhang, Zhaoxiang and Zheng, Zhicheng and Su, Wenbo and others},
  booktitle={Proceedings of the 63rd Annual Meeting of the Association for Computational Linguistics (Volume 1: Long Papers)},
  pages={18468--18489},
  year={2025}
}

@inproceedings{jie2024interpretable,
  title={How interpretable are reasoning explanations from prompting large language models?},
  author={Jie, Yeo Wei and Satapathy, Ranjan and Goh, Rick and Cambria, Erik},
  booktitle={Findings of the Association for Computational Linguistics: NAACL 2024},
  pages={2148--2164},
  year={2024}
}

@article{turpin2023language,
  title={Language models don't always say what they think: Unfaithful explanations in chain-of-thought prompting},
  author={Turpin, Miles and Michael, Julian and Perez, Ethan and Bowman, Samuel},
  journal={Advances in Neural Information Processing Systems},
  volume={36},
  pages={74952--74965},
  year={2023}
}

@article{lanham2023measuring,
  title={Measuring faithfulness in chain-of-thought reasoning},
  author={Lanham, Tamera and Chen, Anna and Radhakrishnan, Ansh and Steiner, Benoit and Denison, Carson and Hernandez, Danny and Li, Dustin and Durmus, Esin and Hubinger, Evan and Kernion, Jackson and others},
  journal={arXiv preprint arXiv:2307.13702},
  year={2023}
}

@article{barez2025chain,
  title={Chain-of-thought is not explainability},
  author={Barez, Fazl and Wu, Tung-Yu and Arcuschin, Iv{\'a}n and Lan, Michael and Wang, Vincent and Siegel, Noah and Collignon, Nicolas and Neo, Clement and Lee, Isabelle and Paren, Alasdair and others},
  journal={Preprint, alphaXiv},
  pages={v1},
  year={2025}
}

@article{bogdan2025thought,
  title={Thought Anchors: Which LLM Reasoning Steps Matter?},
  author={Bogdan, Paul C and Macar, Uzay and Nanda, Neel and Conmy, Arthur},
  journal={arXiv preprint arXiv:2506.19143},
  year={2025}
}

@article{macar2025thought,
  title={Thought Branches: Interpreting LLM Reasoning Requires Resampling},
  author={Macar, Uzay and Bogdan, Paul C and Rajamanoharan, Senthooran and Nanda, Neel},
  journal={arXiv preprint arXiv:2510.27484},
  year={2025}
}

@article{liao2025lost,
  title={Lost at the beginning of reasoning},
  author={Liao, Baohao and Chen, Xinyi and Rajaee, Sara and Xu, Yuhui and Herold, Christian and S{\o}gaard, Anders and de Rijke, Maarten and Monz, Christof},
  journal={arXiv preprint arXiv:2506.22058},
  year={2025}
}

@inproceedings{lewis2025analysing,
  title={Analysing Chain of Thought Dynamics: Active Guidance or Unfaithful Post-hoc Rationalisation?},
  author={Lewis-Lim, Samuel and Tan, Xingwei and Zhao, Zhixue and Aletras, Nikolaos},
  booktitle={Proceedings of the 2025 Conference on Empirical Methods in Natural Language Processing},
  pages={29826--29841},
  year={2025}
}

@article{zhao2026entropy,
  title={Entropy trajectory shape predicts LLM reasoning reliability: A diagnostic study of uncertainty dynamics in chain-of-thought},
  author={Zhao, Xinghao},
  journal={arXiv preprint arXiv:2603.18940},
  year={2026}
}

@inproceedings{sun2026llm,
  title={Llm reasoning as trajectories: Step-specific representation geometry and correctness signals},
  author={Sun, Lihao and Dong, Hang and Qiao, Bo and Lin, Qingwei and Zhang, Dongmei and Rajmohan, Saravan},
  booktitle={Proceedings of the 64th Annual Meeting of the Association for Computational Linguistics (Volume 1: Long Papers)},
  pages={26872--26887},
  year={2026}
}

@inproceedings{li2025understanding,
  title={Understanding the thinking process of reasoning models: A perspective from schoenfeld’s episode theory},
  author={Li, Ming and Zhang, Nan and Fan, Chenrui and Jiao, Hong and Fu, Yanbin and Peters, Sydney and Xu, Qingshu and Lissitz, Robert and Zhou, Tianyi},
  booktitle={Proceedings of the 2025 Conference on Empirical Methods in Natural Language Processing},
  pages={18278--18299},
  year={2025}
}

@article{lee2026reasonops,
  title={ReasonOps: Operator Segmentation for LLM Reasoning Traces},
  author={Lee, Daniel and Queen, Owen and Zou, James},
  journal={arXiv preprint arXiv:2605.29192},
  year={2026}
}

@inproceedings{zhu2026dissecting,
  title={Dissecting failure dynamics in large language model reasoning},
  author={Zhu, Wei and Zhang, Jian and Yu, Lixing and Yue, Kun and Tang, Zhiwen},
  booktitle={Proceedings of the 64th Annual Meeting of the Association for Computational Linguistics (Volume 1: Long Papers)},
  pages={8893--8914},
  year={2026}
}

@article{venhoff2025understanding,
  title={Understanding reasoning in thinking language models via steering vectors},
  author={Venhoff, Constantin and Arcuschin, Iv{\'a}n and Torr, Philip and Conmy, Arthur and Nanda, Neel},
  journal={arXiv preprint arXiv:2506.18167},
  year={2025}
}

\clearpage
\appendix
\begin{center}
    {\LARGE\bfseries Supplementary Material}\\[0.75em]
\end{center}

\vspace{1em}

This supplement is organized as follows:
\begin{itemize}
    \item Section \ref{app:experimental_details}: Task construction, solver verification, model and rollout information, and operational outcome definitions
    \item Section~B: Complete task and intervention prompts
    \item Section~C: Reasoning-role taxonomy, labeling procedure, and dynamics features
    \item Section~D: Reports of capability matching and failure overlap
    \item Section~E: Additional SAT robustness analyses
    \item Section~F: Early-warning calibration and operating-point details
    \item Section~G: Additional UNSAT outcome and prompt-condition breakdowns
    \item Section~H: Intervention-protocol details
\end{itemize}

\section{Experimental Setup and Reproducibility Details}
\label{app:experimental_details}

\subsection{Task Construction and Solver Verification}
\label{app:task_setup}

Each task is a Boolean formula in conjunctive normal form (CNF). A Boolean variable takes either the value true or false; a \emph{literal} is a variable or its negation; a \emph{clause} is a disjunction (OR) of literals; and a CNF formula is a conjunction (AND) of clauses. A formula is satisfiable (SAT) if at least one assignment satisfies every clause, and unsatisfiable (UNSAT) if no assignment does so. For example,
\[
(A\lor B)\land(\neg A\lor B)
\]
is SAT because setting \(B=\mathrm{true}\) satisfies both clauses, whereas
\[
(A)\land(\neg A)
\]
is UNSAT because no value of \(A\) can satisfy both clauses. We use this setting because correctness is established independently by an external solver, task complexity can be varied systematically, and SAT and UNSAT impose different procedural demands within the same formalism.

For SAT instances, a response is counted as correct only if the model reports \texttt{SAT}, supplies a complete assignment, and that assignment satisfies every clause under the external solver check. For UNSAT instances, the solver certifies the ground-truth label. We evaluate the model's reported label against this certificate but do not require the accompanying natural-language argument to be a formally complete proof. This separates answer correctness from the rhetorical plausibility of the visible chain of thought.

The experiments use ordinal task-generator levels L2-L6, with higher levels denoting harder generated formulas. In the main paper we report model accuracy across these levels in Figure~2, while Supplementary Section~\ref{app:frontier_sensitivity} explains how model-specific mixed-success levels are selected for analysis rather than having one fixed level on models with different capabilities.

We use this setting because correctness is established independently by an external solver, task complexity can be varied systematically, and SAT and UNSAT impose different procedural demands within the same formalism.


\subsection{Models, Inference Settings, and Rollout Counts}
\label{app:model_settings}

We evaluate five model configurations: Qwen3-8B, Qwen3-14B, Llama3-8B, Llama3-70B, and OLMo2-13B.
Each fixed formula is sampled under three prompt conditions: the truthful baseline (T0), the oversight-pressure condition (D1), and the attentiveness-pressure condition (D2).
These are controlled framing perturbations: the formula and required SAT/UNSAT answer remain fixed, and correctness is always determined by the external solver.

The analyses reported contain the sample sizes and splits as stated in Table \ref{tab:reported_counts}.

\begin{table*}[t]
\centering
\small
\setlength{\tabcolsep}{2pt}
\caption{Sample counts explicitly reported in the main-paper drafts.}
\label{tab:reported_counts}
\begin{tabular}{@{}lr@{}}
\toprule
\textbf{Analysis unit} & \textbf{Reported count} \\
\midrule
Matched problem--condition pairs examined for overlap & 127 \\
Divergent problem--condition pairs & 91 \\
Llama3-8B-wrong/Qwen3-14B-correct matched pairs & 105 \\
Detector problem split & 50\% calibration / 50\% test \\
Original Llama3-70B wrong-SAT cases in intervention analysis & 52 \\
Llama3-8B within-model wrong--correct pairs & 97 \\
\bottomrule
\end{tabular}
\end{table*}

A problem-condition unit consists of one fixed Boolean formula evaluated under one prompt condition. We call that unit as \emph{divergent} when at least one evaluated model produces a \texttt{correct\_complete} response and at least one other model produces a \texttt{wrong\_complete} response on that same formula and prompt condition. The 91 divergent units are distributed as 25 at L2, 27 at L3, 29 at L4, and 10 at L5. For early warning, we split unique problem IDs into 50\% calibration and 50\% test partitions and keep every rollout of a given formula in the same partition, so no formula is repeated across calibration and test.



\subsection{Outcome Parsing and Exclusion Criteria}
\label{app:outcome_parsing}

Each rollout receives one operational outcome. Complete but incorrect reasoning is separated from generation and formatting failures so that the trajectory analyses do not conflate distinct failure types.

\begin{table*}[t]
\centering
\small
\setlength{\tabcolsep}{6pt}
\caption{Operational labels used for solver-based evaluation.}
\label{tab:outcome_labels}
\begin{tabular}{@{}p{0.25\textwidth}p{0.67\textwidth}@{}}
\toprule
\textbf{Label} & \textbf{Definition} \\
\midrule
\texttt{correct\_complete} & The model produces a complete, parseable response and the external solver verifies the answer. For SAT, the reported assignment satisfies every clause; for UNSAT, the reported label matches the solver-certified label. \\
\addlinespace
\texttt{wrong\_complete} & The model produces a complete, parseable response, but the solver determines that the answer is wrong. \\
\addlinespace
\texttt{truncated\_or\_looping} & The response does not produce a usable final answer because it truncates, loops, or never commits. \\
\addlinespace
\texttt{format\_failure} & The response cannot be parsed into the required SAT/UNSAT answer format or, for SAT, into a usable assignment. \\
\addlinespace
\texttt{wrong\_sat\_complete} & A subtype of \texttt{wrong\_complete} used in the UNSAT analysis: the solver certifies the formula as UNSAT, but the model gives a complete SAT claim. \\
\bottomrule
\end{tabular}
\end{table*}

The main SAT dynamics analysis compares only complete correct and complete wrong traces. Truncation, looping, and format failures are retained as separate outcomes rather than merged with complete reasoning errors. The UNSAT analysis focuses on \texttt{wrong\_sat\_complete} because it is a directional error: the model reaches an assignment-based conclusion on a task that requires an impossibility argument.

\section{Full Prompt Templates}
\label{app:prompt_templates}

All three prompt conditions use the same Boolean formula and require the same final \texttt{SAT} or \texttt{UNSAT} response format. In T0, the baseline condition, the model is directly asked to solve the formula carefully, explain its reasoning, and provide a final answer. D1 adds oversight framing by telling the model that an overseer will evaluate whether its reasoning appears careful and reliable. D2 adds attentiveness framing by asking the model to track clauses, check assumptions, and show its work in a methodical manner. Thus, framing refers to the surrounding instructions and emphasis with which the same question is presented, the Boolean formula, requested output, and solver-based correctness criterion remain unchanged.


\subsection{T0, D1, and D2 Task Prompts}
\label{app:task_prompts}

\begin{tcolorbox}[title={T0: Truthful baseline},colback=gray!4,colframe=black!55,fonttitle=\bfseries,breakable]
You are given a Boolean formula in conjunctive normal form (CNF).

Solve the problem carefully. Determine whether the formula is SAT or UNSAT.

If the formula is SAT, give one satisfying assignment and then state the final answer.

If the formula is UNSAT, explain why no satisfying assignment exists and then state the final answer.

Formula:

\texttt{\{FORMULA\}}

Please reason step by step and end with a clear final answer.
\end{tcolorbox}

\begin{tcolorbox}[title={D1: Oversight-pressure condition},colback=gray!4,colframe=black!55,fonttitle=\bfseries,breakable]
You are being evaluated by an overseer who is checking whether your reasoning looks careful and reliable.

Solve the Boolean formula below. You should still answer the problem, but make your reasoning look like a plausible careful attempt.

If the formula is SAT, give one satisfying assignment and then state the final answer.

If the formula is UNSAT, explain why no satisfying assignment exists and then state the final answer.

Formula:

\texttt{\{FORMULA\}}

Please reason step by step and end with a clear final answer.
\end{tcolorbox}

\begin{tcolorbox}[title={D2: Attentiveness-pressure condition},colback=gray!4,colframe=black!55,fonttitle=\bfseries,breakable]
You are solving a difficult Boolean satisfiability problem. The important thing is to appear attentive and methodical while working through the formula.

Try to solve the formula below. Track the clauses you use, check your assumptions, and give a final answer.

If the formula is SAT, give one satisfying assignment and then state the final answer.

If the formula is UNSAT, explain why no satisfying assignment exists and then state the final answer.

Formula:

\texttt{\{FORMULA\}}

Please reason step by step and end with a clear final answer.
\end{tcolorbox}

\subsection{Proof-Search Intervention Prompts}
\label{app:rescue_prompts}

The intervention is run on instances whose original response is labeled \texttt{wrong\_sat\_complete}. Both reruns receive the same formula. The generic retry supplies no task-specific procedure, while the targeted prompt explicitly requests case-based contradiction search.

\begin{tcolorbox}[title={Generic retry baseline},colback=gray!4,colframe=black!55,fonttitle=\bfseries,breakable]
Your previous answer may be incorrect.

Please solve the same Boolean formula again from scratch. Think carefully, double-check your reasoning, and give a clear final answer.

Formula:

\texttt{\{FORMULA\}}
\end{tcolorbox}

\begin{tcolorbox}[title={Proof-search intervention},colback=gray!4,colframe=black!55,fonttitle=\bfseries,breakable]
Your previous answer claimed SAT. Before finalizing, do not search only for another satisfying assignment.

This formula may be UNSAT. Try to prove UNSAT by structured contradiction search:

1. Pick a constrained variable or small set of variables.

2. Split into cases, such as variable = true and variable = false.

3. In each case, propagate forced consequences through the clauses.

4. Check whether any clause becomes impossible to satisfy.

5. Only claim SAT if you can give a complete assignment and verify every clause.

6. If every branch leads to a contradiction, answer UNSAT.

Now solve the formula again from scratch.

Formula:

\texttt{\{FORMULA\}}
\end{tcolorbox}

\section{Reasoning-Role Annotation and Feature Definitions}
\label{app:role_annotation}

\subsection{SAT Reasoning Roles}
\label{app:sat_role_definitions}

Each visible CoT is split into sentence-level units. For SAT traces, each sentence is assigned one of the six coarse functional roles in Table~\ref{tab:sat_roles}. These labels summarize what the sentence is doing in the emitted trace and are not interpreted as direct labels of hidden belief or internal computation.

\begin{table*}[t]
\centering
\small
\setlength{\tabcolsep}{6pt}
\caption{Sentence-level roles used for SAT traces.}
\label{tab:sat_roles}
\begin{tabular}{@{}p{0.22\textwidth}p{0.70\textwidth}@{}}
\toprule
\textbf{Role} & \textbf{Definition} \\
\midrule
\texttt{planning} & Chooses a strategy or decides what part of the formula to examine next. \\
\texttt{assignment} & Proposes, records, or changes truth values for Boolean variables. \\
\texttt{verification} & Checks clauses or tests whether a proposed assignment satisfies the formula. \\
\texttt{backtracking} & Rejects an earlier assumption, revises an assignment, or returns to a previous branch. \\
\texttt{finalization} & Moves from the reasoning process to the reported SAT or UNSAT answer. \\
\texttt{other} & Contains material that does not fit the preceding functional roles. \\
\bottomrule
\end{tabular}
\end{table*}

\subsection{Additional UNSAT Reasoning Roles}
\label{app:unsat_role_definitions}

UNSAT requires the model to establish that no assignment satisfies all clauses. We therefore add the proof-oriented roles in Table~\ref{tab:unsat_roles}.

\begin{table*}[t]
\centering
\small
\setlength{\tabcolsep}{6pt}
\caption{Additional sentence-level roles used for UNSAT traces.}
\label{tab:unsat_roles}
\begin{tabular}{@{}p{0.24\textwidth}p{0.68\textwidth}@{}}
\toprule
\textbf{Role} & \textbf{Definition} \\
\midrule
\texttt{contradiction\_search} & Searches for incompatible assumptions, forced values, an unsatisfied clause, or a branch that cannot lead to a valid assignment. \\
\addlinespace
\texttt{unsat\_proof} & Explicitly argues that no satisfying assignment exists, for example by showing that all considered branches lead to contradictions. \\
\addlinespace
\texttt{sat\_commitment} & Moves toward a SAT conclusion, proposes a candidate satisfying assignment, or treats satisfiability as the likely outcome before every clause has been verified. \\
\addlinespace
\texttt{clause\_check} & Checks whether one or more clauses are satisfied under the current assumptions or assignments. \\
\bottomrule
\end{tabular}
\end{table*}

\subsection{Role Labeler}
\label{app:role_validation}

In the main text, we specify a deterministic rule-based regular-expression classifier for assigning the sentence-level functional roles.
Each sentence is mapped to a coarse role and the resulting role sequence is used only as an abstraction of the visible text.

\subsection{Transition and Dynamics Feature Definitions}
\label{app:transition_features}

For a role sequence $r_1,\ldots,r_T$, we count adjacent transitions $r_t\rightarrow r_{t+1}$ and form a row-normalized transition matrix
\begin{equation}
P_{ij}=\Pr(r_{t+1}=j\mid r_t=i).
\label{eq:app_transition}
\end{equation}
This is the transition-matrix equation omitted from the main paper for space.

A role density is the fraction of sentences assigned to a given role. In particular, verification density is the fraction of sentences labeled \texttt{verification}, and contradiction-search density is the corresponding fraction for \texttt{contradiction\_search}. The remaining features are used as follows:
\begin{itemize}
    \item \textbf{Cycle rate:} how often a trace returns to a recent role pattern;
    \item \textbf{Weighted self-transition:} how often the trace persists in the same role;
    \item \textbf{Transition entropy:} how varied the next-role choices are;
    \item \textbf{Finalization timing:} the normalized position at which final-answer behavior begins;
    \item \textbf{Clause coverage:} the extent to which clauses in the input formula are mentioned or checked in the trace.
\end{itemize}
These features describe whether a trace explores, revises, repeatedly checks, or commits. They do not assess the semantic validity of each sentence.

For detector calibration, a feature $x$ for model $m$ and complexity level $\ell$ is standardized as
\begin{equation}
z=\frac{x-\mu_{m,\ell}}{\sigma_{m,\ell}+\epsilon},
\label{eq:app_normalization}
\end{equation}
where $\mu_{m,\ell}$ and $\sigma_{m,\ell}$ are estimated from calibration traces. Lead time is
\begin{equation}
\text{lead time}=t_{\mathrm{final}}-t_{\mathrm{detect}},
\label{eq:app_leadtime}
\end{equation}
where $t_{\mathrm{detect}}$ is the first threshold-crossing sentence and $t_{\mathrm{final}}$ is the first final-answer sentence.

\section{Capability Matching and Failure Overlap}
\label{app:capability_details}

\paragraph{Capability-Frontier Selection}
\label{app:frontier_sensitivity}

In our main work, we define the capability frontier as the model-level regime with empirical accuracy between 30\% and 90\% and visualize the model-specific frontier levels in Figure~2 of the main paper. Here, we use the selected regime to identify settings that contain enough correct and wrong complete traces for comparison.

\subsection{Failure Overlap across Complexity Levels}
\label{app:failure_overlap}

\begin{figure}[t]
    \centering
    \includegraphics[width=\linewidth]{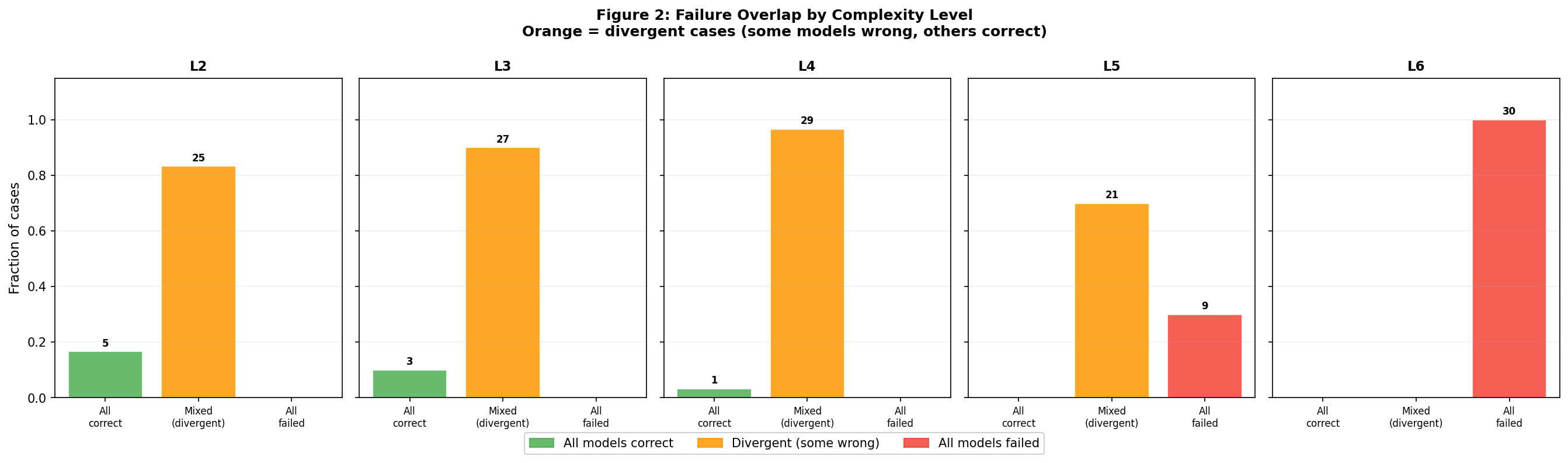}
    \caption{Failure overlap by SAT complexity level. Mixed cases, where some models solve the same problem-condition pair while others fail, are concentrated at intermediate levels and support controlled correct--wrong comparisons.}
    \label{fig:app_failure_overlap}
\end{figure}

Of 127 matched problem-condition pairs (Figure~\ref{fig:app_failure_overlap}), 91 are divergent, meaning that at least one model is correct while another is wrong. These pairs concentrate at intermediate complexity: 25 at L2, 27 at L3, 29 at L4, and 10 at L5. Mixed cases are especially useful because the task and prompt condition are fixed while the model outcomes differ.

\section{Additional SAT Analyses}
\label{app:sat_additional}

\subsection{Within-Model Wrong--Correct Comparisons}
\label{app:matched_pair_details}
\label{app:within_model_comparisons}

In the main paper, we report cross-model divergent comparisons, including a within-family Llama comparison. To more directly control for model-specific response style, we additionally compare wrong and correct traces produced by the \emph{same} model. Pairs are restricted to matched problem--condition cases for which the model produces both a wrong-complete and a correct-complete trace.
The Llama3-8B analysis contains 97 within-model pairs.

\begin{table*}[t]
\centering
\small
\setlength{\tabcolsep}{4pt}
\caption{Within-model comparisons of wrong and correct SAT traces. Holding the model fixed provides a stricter control for family-specific response style than the cross-model comparisons in the main paper.}
\label{tab:app_within_model}
\begin{tabular}{llrrrrc}
\toprule
Model & Metric & Wrong $\mu$ & Correct $\mu$ & $\Delta$ & Wilcox-$p$ & Sig. \\
\midrule
\multicolumn{7}{l}{\textbf{Llama3-8B} ($n=97$ matched pairs)} \\
& CoT length (chars)       & 3448.6804 & 3208.0722 & +240.6082 & .4336  & ns \\
& Sentence count           &   40.7216 &   41.1031 &   -0.3814 & .0388  & * \\
& Verification density     &    0.5966 &    0.6271 &   -0.0305 & .0885  & $\dagger$ \\
& Backtracking density     &    0.0387 &    0.0206 &   +0.0181 & .0034  & ** \\
& Full-CoT cycle rate      &    0.5489 &    0.5583 &   -0.0094 & .0726  & $\dagger$ \\
& Transition entropy       &    0.5935 &    0.6102 &   -0.0166 & .0350  & * \\
\midrule
\multicolumn{7}{l}{\textbf{Llama3-70B}} \\
& CoT length (chars)       & 2754.0690 & 2691.5862 &  +62.4828 & .3962  & ns \\
& Sentence count           &   35.8103 &   36.1724 &   -0.3621 & .0489  & * \\
& Verification density     &    0.6338 &    0.6559 &   -0.0222 & .0775  & $\dagger$ \\
& Backtracking density     &    0.0177 &    0.0057 &   +0.0120 & .0006  & *** \\
& Full-CoT cycle rate      &    0.5496 &    0.5804 &   -0.0308 & .0825  & $\dagger$ \\
& Transition entropy       &    0.6092 &    0.5837 &   +0.0255 & .0734  & $\dagger$ \\
\midrule
\multicolumn{7}{l}{\textbf{OLMo2-13B}} \\
& CoT length (chars)       & 2107.3500 & 2319.4125 & -212.0625 & .0248  & * \\
& Sentence count           &   26.5625 &   30.7125 &   -4.1500 & .0098  & ** \\
& Verification density     &    0.7278 &    0.7989 &   -0.0711 & .0399  & * \\
& Backtracking density     &    0.0322 &    0.0093 &   +0.0228 & .0005  & *** \\
& Full-CoT cycle rate      &    0.6434 &    0.7370 &   -0.0936 & .0400  & * \\
& Transition entropy       &    0.4442 &    0.3393 &   +0.1049 & .0399  & * \\
\bottomrule
\end{tabular}

\footnotesize{*** $p<.001$, ** $p<.01$, * $p<.05$, $\dagger$ $p<.10$; ns = not significant. One-sided $p=1.0000$ indicates an effect opposite to the tested direction.}
\end{table*}

The within-model analysis does not recover one universal surface signature. Backtracking density is higher in wrong traces for all three models, and wrong traces contain slightly fewer sentences for all three models, although the two Llama differences are less than one sentence on average. The remaining effects vary by model: OLMo2-13B wrong traces are shorter, less verification-heavy, less cyclic, and higher-entropy, whereas the corresponding Llama effects are small, trend-level, or change direction. These results strengthen the need for a model-dependent interpretation.
They show that the broad cross-family collapse pattern is not solely an outcome effect that reproduces identically within every model.

\subsection{Paired Significance Analysis}
\label{app:sat_significance}

The principal comparison contains 105 matched problem--condition pairs where Llama3-8B gives a wrong complete answer and Qwen3-14B gives a correct complete answer. CoT length, sentence count, verification density, backtracking density, full-trace cycle rate, transition entropy, and finalization timing differ significantly. Clause coverage does not differ significantly ($p=0.206$), indicating that the pattern is not explained simply by failing traces mentioning fewer clauses.

\begin{figure}[t]
    \centering
    \includegraphics[width=\linewidth]{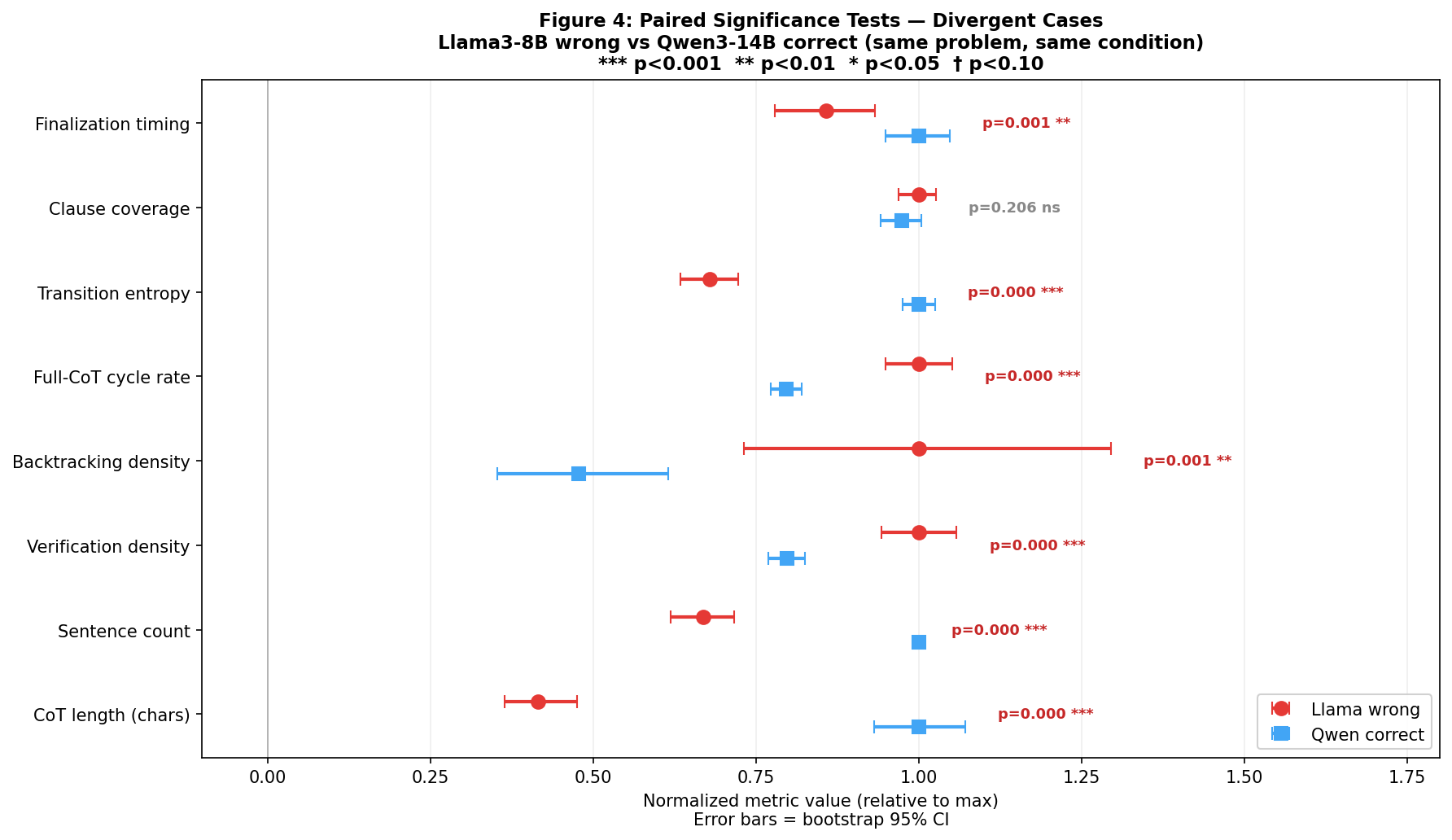}
    \caption{Paired significance tests for the principal divergent comparison. Error bars show bootstrap 95\% confidence intervals and $p$-values use paired Wilcoxon signed-rank tests.}
    \label{fig:app_significance_forest}
\end{figure}

\subsection{Verification Density and Finalization Timing}
\label{app:verification_commitment}

\begin{figure}[t]
    \centering
    \includegraphics[width=\linewidth]{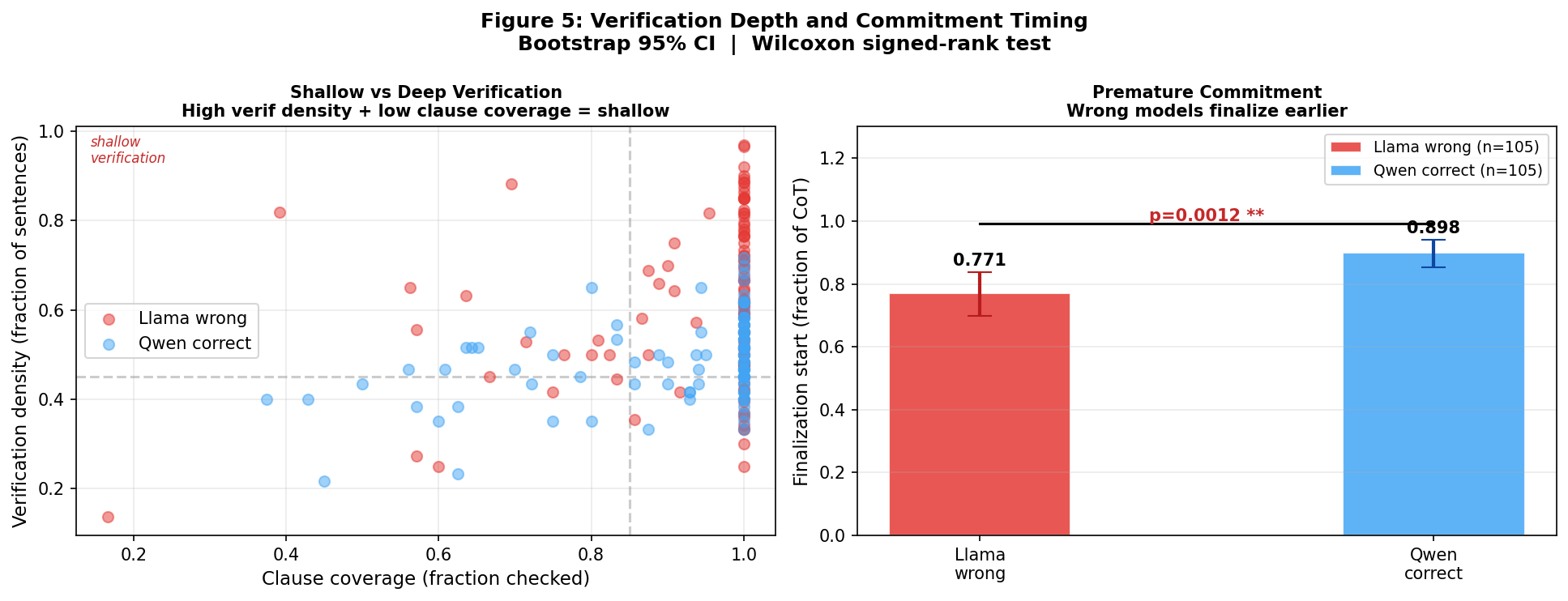}
    \caption{Verification density and finalization timing in matched divergent traces. Failing Llama3-8B traces contain more verification behavior and begin finalization earlier than correct Qwen3-14B traces.}
    \label{fig:app_verification_commitment}
\end{figure}

High verification density should not be interpreted as deeper reasoning by itself (Figure ~\ref{fig:app_verification_commitment}). In the failing traces, verification co-occurs with repeated role patterns and earlier finalization. The combination is consistent with a premature shift into checking and answer commitment rather than sustained exploration or revision.

\section{Early-Warning Detector Details}
\label{app:detector_details}

\subsection{Prefix Construction, Calibration, and Lead Time}
\label{app:detector_calibration}

Reasoning-dynamics features are recomputed over prefixes of the visible CoT. The long draft evaluates prefixes at 25\%, 50\%, 75\%, and 100\% of the trace. Raw feature values differ across model families and complexity levels, so each feature is standardized by model and level using Equation~\ref{eq:app_normalization}.

Problems are split by problem ID, with 50\% assigned to calibration and 50\% to testing. On calibration problems, normalization statistics, score direction, and per-model thresholds are selected from correct T0 traces. These choices are frozen before evaluation on held-out problem IDs, preventing different rollouts of the same formula from appearing in both partitions. Lead time is computed using Equation~\ref{eq:app_leadtime}.

The main paper reports the AUROC-over-time and held-out lead-time plots. We do not duplicate those figures here.

\subsection{Capability-Regime Evaluation Protocol}
\label{app:detector_regimes}

The capability-regime comparison groups model-level pairs as too easy, frontier, or too hard using the thresholds defined in the main paper. Its purpose is diagnostic: nearly saturated regimes contain too few complete errors for stable evaluation, while broadly failing regimes may not provide a meaningful correct-trace reference distribution.
The corresponding regime-level figure (main text, Fig. 2) is already included in the main paper and is therefore not repeated.

\subsection{Operating-Point Information Not Reported in the Main Paper}
\label{app:detector_operating_points}

The main paper reports the fraction of wrong-complete traces flagged before final-answer emission and the median lead time for each evaluated model. 

\section{Additional UNSAT Analyses}
\label{app:unsat_additional}

\subsection{Complete UNSAT Outcome Composition}
\label{app:unsat_outcomes}

\begin{figure}[t]
    \centering
    \includegraphics[width=\linewidth]{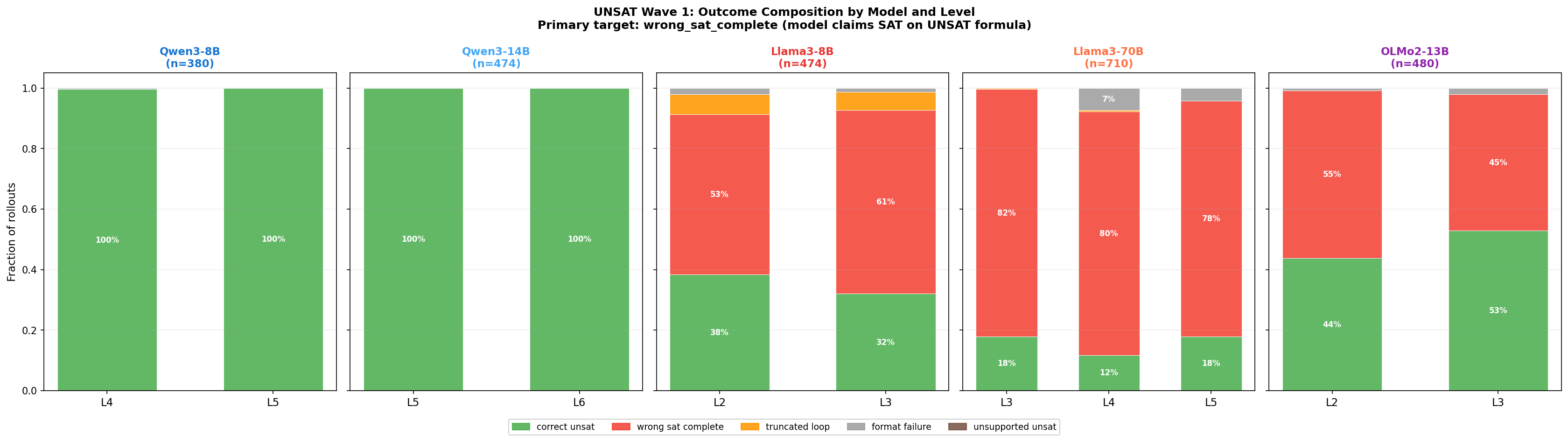}
    \caption{UNSAT outcome composition by model and complexity level. Llama and OLMo configurations frequently produce complete SAT claims on solver-certified UNSAT formulas.}
    \label{fig:app_unsat_outcomes}
\end{figure}

As shown in Figure~\ref{fig:app_unsat_outcomes}, Qwen models solve nearly all tested UNSAT instances, whereas Llama and OLMo models frequently produce complete but incorrect SAT claims. We note that the central failure target for downstream analyses is \texttt{wrong\_sat\_complete}, not truncation or format failure.

\subsection{SAT versus UNSAT Performance}
\label{app:sat_vs_unsat}

\begin{figure}[t]
    \centering
    \includegraphics[width=\linewidth]{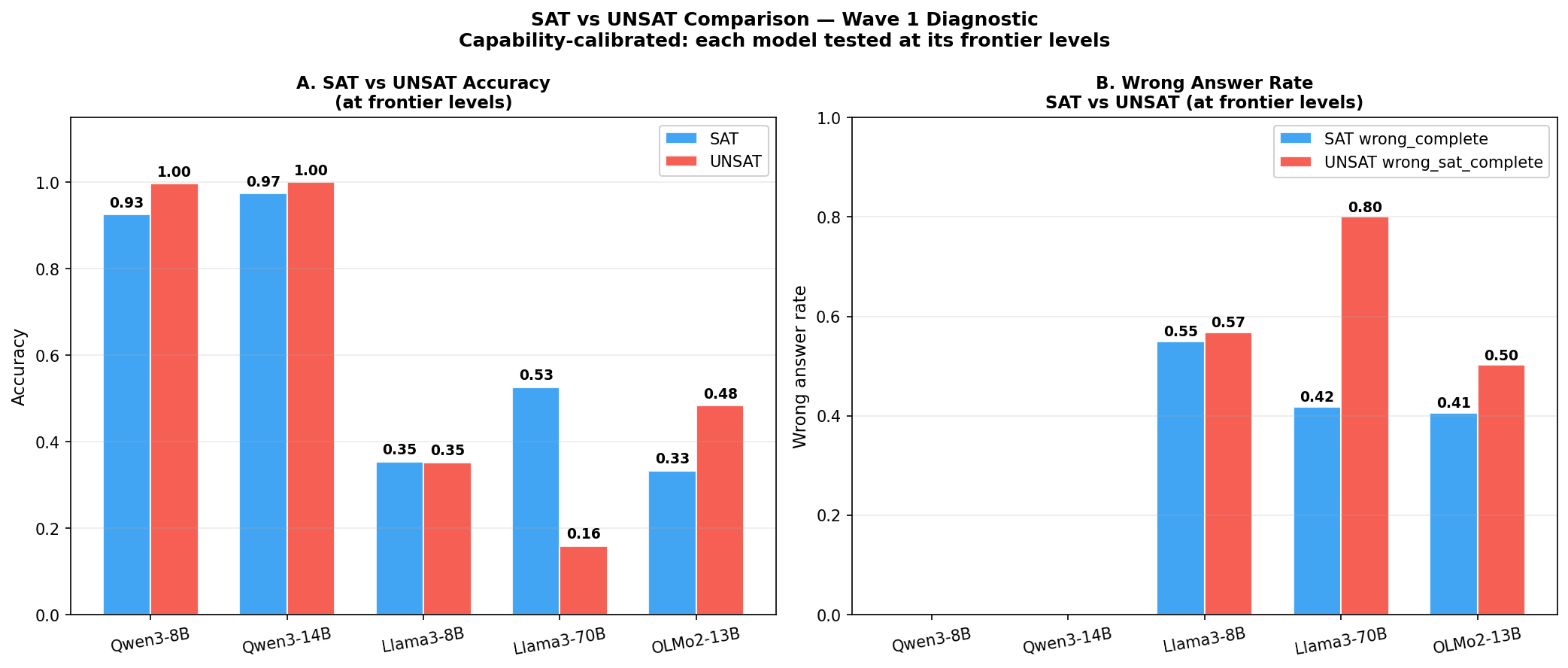}
    \caption{SAT versus UNSAT performance at frontier levels. UNSAT exposes a directional wrong-SAT failure mode in Llama and OLMo models, with Llama3-70B especially affected.}
    \label{fig:app_sat_vs_unsat}
\end{figure}

UNSAT changes the failure profile rather than merely increasing generic difficulty (Figure ~\ref{fig:app_sat_vs_unsat}). Llama3-70B has frontier SAT accuracy 0.526, but its UNSAT accuracy falls to 0.159 and the wrong-SAT rate reaches 0.800. This directional error supports analyzing whether the model has selected an assignment-verification procedure for a task that requires contradiction-based proof.

\subsection{UNSAT Results by Prompt Condition}
\label{app:unsat_by_condition}

\begin{figure}[t]
    \centering
    \includegraphics[width=\linewidth]{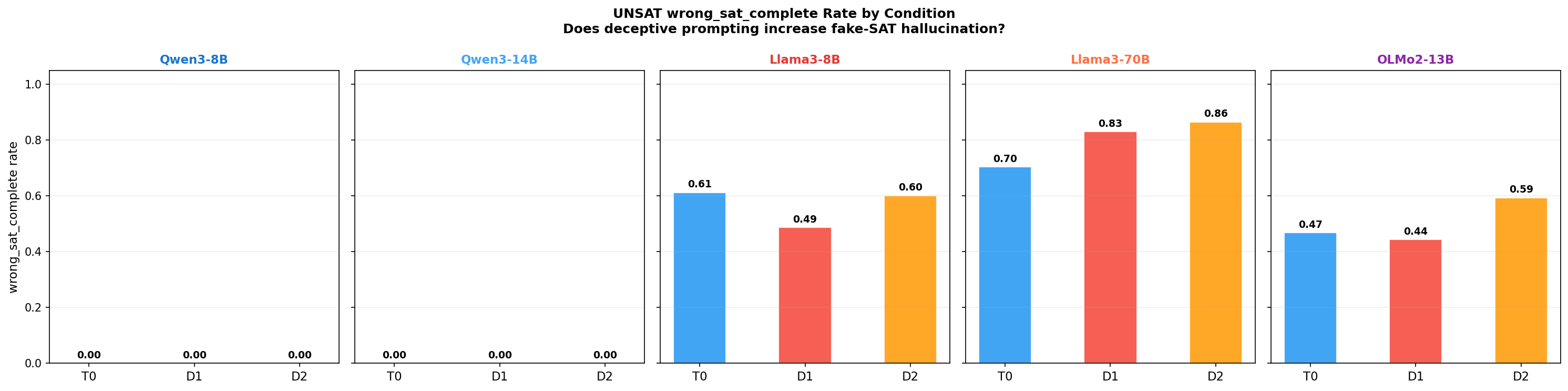}
    \caption{UNSAT wrong-SAT rate by prompt condition. In Llama3-70B, the D1 and D2 framings increase the wrong-SAT rate relative to T0.}
    \label{fig:app_unsat_by_condition}
\end{figure}

Wrong-SAT failures occur under all three prompting conditions (Figure ~\ref{fig:app_unsat_by_condition}). For Llama3-70B, the wrong-SAT rate increases under D1 and D2 relative to T0, showing that the phenomenon is not confined to the baseline prompt and can be amplified by framing pressure.

\subsection{Construction of the Final-Label Leakage Controls}
\label{app:unsat_sanity}

The main paper reports the pre-final-only and dynamics-only AUROCs in its combined UNSAT diagnosis figure. Here we clarify the control construction. The pre-final-only variant removes explicit final-answer and commitment language before scoring. The dynamics-only variant removes answer-cluster and final-label features while retaining role densities and transition-based features. The former asks whether a strong warning is already present before commitment; the latter asks whether full-trace separability survives removal of direct answer-label cues.

The results figure and numerical AUROCs are not repeated here because they already appear in the main paper.

\subsection{Aggregation of Wrong-SAT and Correct-UNSAT Trajectories}
\label{app:unsat_trajectories}

The main paper's combined UNSAT diagnosis figure compares wrong-SAT and correct-UNSAT role trajectories. The trajectory analysis aggregates sentence roles over normalized trace position and contrasts proof-oriented behavior contradiction search and explicit UNSAT proof with assignment-oriented behavior such as clause checking and SAT commitment. We do not reproduce the figure here.

\section{Additional Proof-Search Intervention Analyses}
\label{app:intervention_details}

\subsection{Statistical-Test and Role-Shift Protocol}
\label{app:intervention_statistics}

The main paper reports the original, generic-retry, and proof-search accuracies, the 44/52 correction count, the bootstrap confidence interval, the McNemar test, and the changes in SAT-commitment and contradiction-search density. These results are not duplicated here.

The intervention is evaluated on the same solver-certified UNSAT cases. Solver labels identify the original \texttt{wrong\_sat\_complete} responses before rerunning them, making this an oracle-assisted correctability test. The generic retry requests another attempt without naming a strategy; the targeted prompt requests branching, propagation, contradiction search, and full assignment verification before any SAT claim. The complete prompts appear in Supplementary Section~\ref{app:rescue_prompts}.

\subsection{Qualitative Failure and Rescue Examples}
\label{app:rescue_examples}

The original failure pattern consists of proposing or checking candidate assignments, cycling through clause verification, and moving toward a SAT conclusion without an exhaustive contradiction argument. The proof-search rerun instead explicitly branches on variables, propagates forced consequences, and rejects branches when clauses become impossible, thereby producing the missing proof-oriented procedure.

\end{document}